\documentclass[envcountsame]{llncs}
\usepackage{graphicx}
\usepackage{caption}
\usepackage[linesnumbered,lined,ruled,commentsnumbered]{algorithm2e}
\usepackage{comment}
\usepackage[pdfpagelabels,hypertexnames=false,breaklinks=true,bookmarksopen=true,bookmarksopenlevel=2]{hyperref}
\usepackage{setspace}
\usepackage{enumitem}
\usepackage{amsmath}
\usepackage[utf8x]{inputenc}
\usepackage{cite} 
\usepackage{courier}
\usepackage{listings} 
\usepackage[none]{hyphenat}

\usepackage{wasysym} 
\usepackage{multirow,hhline}
\usepackage[table]{xcolor}
\usepackage{dsfont} 
\usepackage{array} 
\newcolumntype{?}{!{\vrule width 1pt}} 
\usepackage{bm} 
\usepackage{makecell} 

\usepackage{tikz,xcolor,hyperref}

\definecolor{lime}{HTML}{A6CE39}
\DeclareRobustCommand{\orcidicon}{%
	\begin{tikzpicture}
	\draw[lime, fill=lime] (0,0) 
	circle [radius=0.16] 
	node[white] {{\fontfamily{qag}\selectfont \tiny ID}};
	\draw[white, fill=white] (-0.0625,0.095) 
	circle [radius=0.007];
	\end{tikzpicture}
	\hspace{-2mm}
}

\foreach \x in {A, ..., Z}{%
	\expandafter\xdef\csname orcid\x\endcsname{\noexpand\href{https://orcid.org/\csname orcidauthor\x\endcsname}{\noexpand\orcidicon}}
}

\begin{document}

\mainmatter  

\title{Towards Building Knowledge by Merging Multiple Ontologies
	with $\mathcal{C}$\textit{o}$\mathcal{M}$\textit{erger}: A Partitioning-based Approach}

\author{Samira Babalou\orcidA{}, Birgitta K\"{o}nig-Ries\orcidC{}}

\institute{Heinz-Nixdorf Chair for Distributed Information Systems \\
	Institute for Computer Science, Friedrich Schiller University Jena, Germany \\
	\email{\{samira.babalou,birgitta.koenig-ries\}@uni-jena.de}}

\maketitle

\begin{sloppypar}

\begin{abstract} 
Ontologies are the prime way of organizing data in the Semantic Web. Often, it is necessary to combine several, independently developed ontologies to obtain a knowledge graph fully representing a domain of interest. The complementarity of existing ontologies can be leveraged by merging them. Existing approaches for ontology merging mostly implement a binary merge. However, with the growing number and size of relevant ontologies across domains, scalability becomes a central challenge. A multi-ontology merging technique offers a potential solution to this problem. We present $\mathcal{C}$\textit{o}$\mathcal{M}$\textit{erger}, a scalable multiple ontologies merging method. For efficient processing, rather than successively merging complete ontologies pairwise, we group related concepts across ontologies into partitions and merge first within and then across those partitions. The experimental results on well-known datasets confirm the feasibility of our approach and demonstrate its superiority over binary strategies. A prototypical implementation is freely accessible through a live web portal.
\end{abstract}

\section{Introduction} \label{intro}
Ontologies represent the semantic model of data on the web. For many usecases, individual ontologies cover just a part of the domain of interest or different ontologies exist that model the domain from different viewpoints. In both cases, by merging them into an integrated knowledge graph, their complementarity can be leveraged and unified knowledge of a given domain can be acquired.
Existing ontology merging approaches \cite{OM,creado,HSSM,granualr,atom,OIMSM} 
are limited to merging two ontologies at a time, due to using a \textit{binary} merge strategy. On the contrary, merging $n$ ontologies ($n>2$) in a single step, employing what is called an \textit{n-ary} strategy, has not been extensively studied so far. 
A series of  binary merges can also be applied incrementally  to more than two ontologies, thus merging more than two ontologies. But, this approach is not  sufficiently scalable and viable for a large number of ontologies \cite{rahm2016case}.

\begin{figure}[tbp]
	\includegraphics[width=10cm]{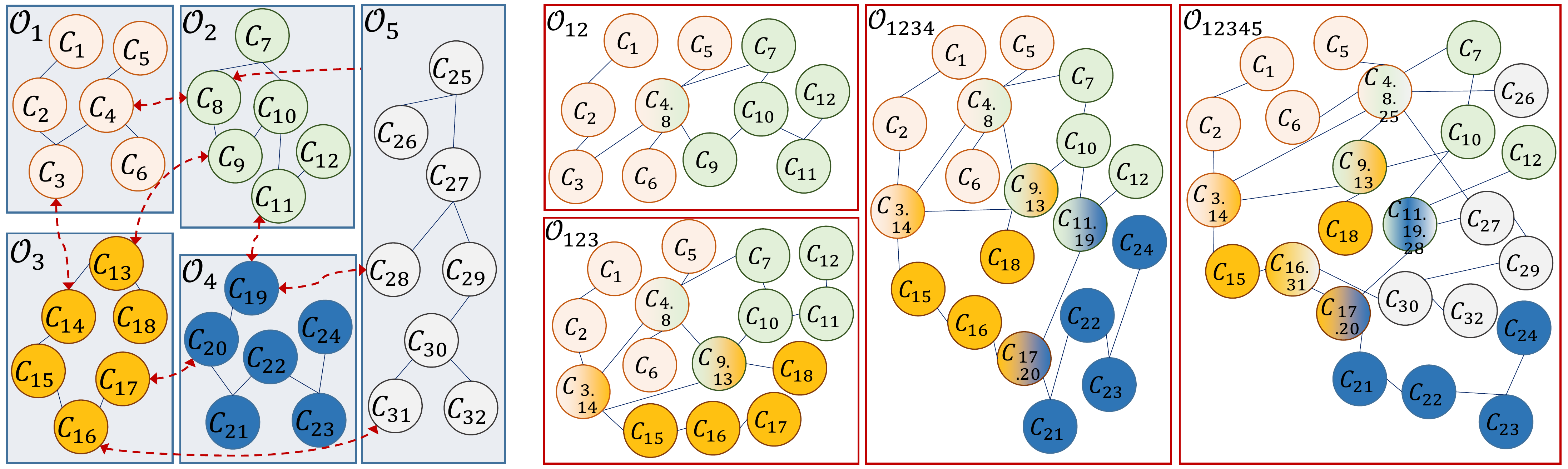}
	\centering
	\caption{Five ontologies with the merged ontologies in each step of binary merge.}
	\label{fig:bn}
	\vspace{-4mm}
\end{figure}
\begin{table}[bt!]
	\caption{Number of operations for merging five sample ontologies.}
	\centering
	\resizebox{7cm}{!} {
		\begin{tabular}{| c | c | c | c | c | c | c |}
			\hline
			\rowcolor{gray!15}
			& \multicolumn{2}{c|}{\textbf{Step}} & \textbf{Source ontologies} & $\boldsymbol{|combine|}$ & $\boldsymbol{|reconst|}$ & $\boldsymbol{|output|}$\\ \hline
			\cellcolor{gray!15}N-ary & \multicolumn{2}{c|}{1} & $\mathcal{O}_1$ \& $\mathcal{O}_2$ \& $\mathcal{O}_3$ \& $\mathcal{O}_4$ \& $\mathcal{O}_5$ & 6 & 28 & 1 \\  \Xhline{3\arrayrulewidth} 
			\cellcolor{gray!15}& \multicolumn{2}{c|}{1} & $\mathcal{O}_1$ \& $\mathcal{O}_2$ & 1 & 5 & 1 \\ \cline{2-7}	
			\cellcolor{gray!15}& \multicolumn{2}{c|}{2} & $\mathcal{O}_{12}$ \& $\mathcal{O}_3$ & 2 & 8 & 1 \\ \cline{2-7}
			\cellcolor{gray!15}Binary & \multicolumn{2}{c|}{3} & $\mathcal{O}_{123}$ \& $\mathcal{O}_4$ & 4 & 6 & 1 \\ \cline{2-7}	
			\cellcolor{gray!15}& \multicolumn{2}{c|}{4} & $\mathcal{O}_{1234}$ \& $\mathcal{O}_5$ & 3 & 13 & 1 \\ \cline{2-7} 
			\cellcolor{gray!15}& Total&4 & $\mathcal{O}_{12345}$ & 10 & 32 & 4 \\ \hline
	\end{tabular}}
	\label{table:Intro-Holistic}
	\vspace{-4mm}%
\end{table}

Let us consider an example to illustrate the differences between the binary and n-ary approaches. Fig.~\ref{fig:bn} shows five source ontologies with their correspondences depicted by dashed lines. To estimate the merge effort, we measure
three operations during merging
: combining the corresponding entities into an integrated entity $|combine|$; reconstructing the relationship $|reconst|$; and output generation 
$|output|$. 
In the binary-ladder strategy~\cite{batini1986comparative}, at the first step, 
$\mathcal{O}_1$ and $\mathcal{O}_2$ are combined into an intermediate merged ontology $\mathcal{O}_{12}$. Then, $\mathcal{O}_{12}$ is merged with $\mathcal{O}_3$ and so on. All intermediate ontologies and final merged ontology are shown in Fig.~\ref{fig:bn} and the required operations are presented in Table~\ref{table:Intro-Holistic}.
The n-ary merged ontology has the same structure as the final merged ontology of the binary-ladder merge for the given source ontologies. 
The binary merge approach needs 10 combinations, while the n-ary method needs 6, which shows 40\% improvement in our example. The number of reconstructions in binary is 28, while n-ary needs 32 operations, which indicates 12.5\% improvement. The n-ary method needs one time output generation, while the binary approach needs 4 times.
While these specific numbers are specific to our example, the general pattern will be the same for other examples. 
The significance of the achieved improvements is impressive especially when dealing with a large number of ontologies and processing large-scale ontologies.

To handle a large number of source ontologies, we aimed to reduce the time and operational complexity while achieving at least the same quality of the final result or even improve upon it.
For efficiently applying the n-ary method on merging multiple ontologies, we utilize a partitioning-based method inspired by partitioning-based ontology matching systems~\cite{hu2008matching,coma,jimenez2018we}. 
These systems first partition the source schemas or ontologies and then perform a partition-wise matching. The idea is to perform partitioning in such a way that every partition of one schema has to be matched with only a subset of the partitions (ideally, only with one partition) of the other schema. This results in a significant reduction of the search space and thus improved efficiency. Furthermore, matching smaller partitions reduces the memory requirements compared to matching the full schemas. 
Following that, in our n-ary method, $\mathcal{C}$\textit{o}$\mathcal{M}$\textit{erger}, we develop an efficient merging technique that scales to many ontologies. 
We show that by using a partitioning-based method, we can reduce the complexity of the search space\footnote{In our context, the search space is the set of entities and their relations that have to be evaluated for a specific merge step.}.
Our method takes as input a set of source ontologies alongside the respective mappings and generates a merged ontology. At first, the $n$ source ontologies are partitioned into $k$ blocks ($k<<n$). 
After that, the blocks are individually merged and refined. Finally, they are combined to produce the merged ontology followed by a global refinement. We provide experimental tests for merging a variety of ontologies showing the effectiveness of our approach over binary approaches.

The rest of the paper is organized as follows: our proposed method is described in Sec. \ref{sec:approach} followed by the experimental results in Sec. \ref{sec:result}. A survey on related work is presented in Sec. \ref{sec:relatedWork} and the paper is concluded in Sec. \ref{sec:conclusion}.

\section{Proposed Method} \label{sec:approach}
\begin{figure}[tbp]
	\includegraphics[width=9cm,keepaspectratio]{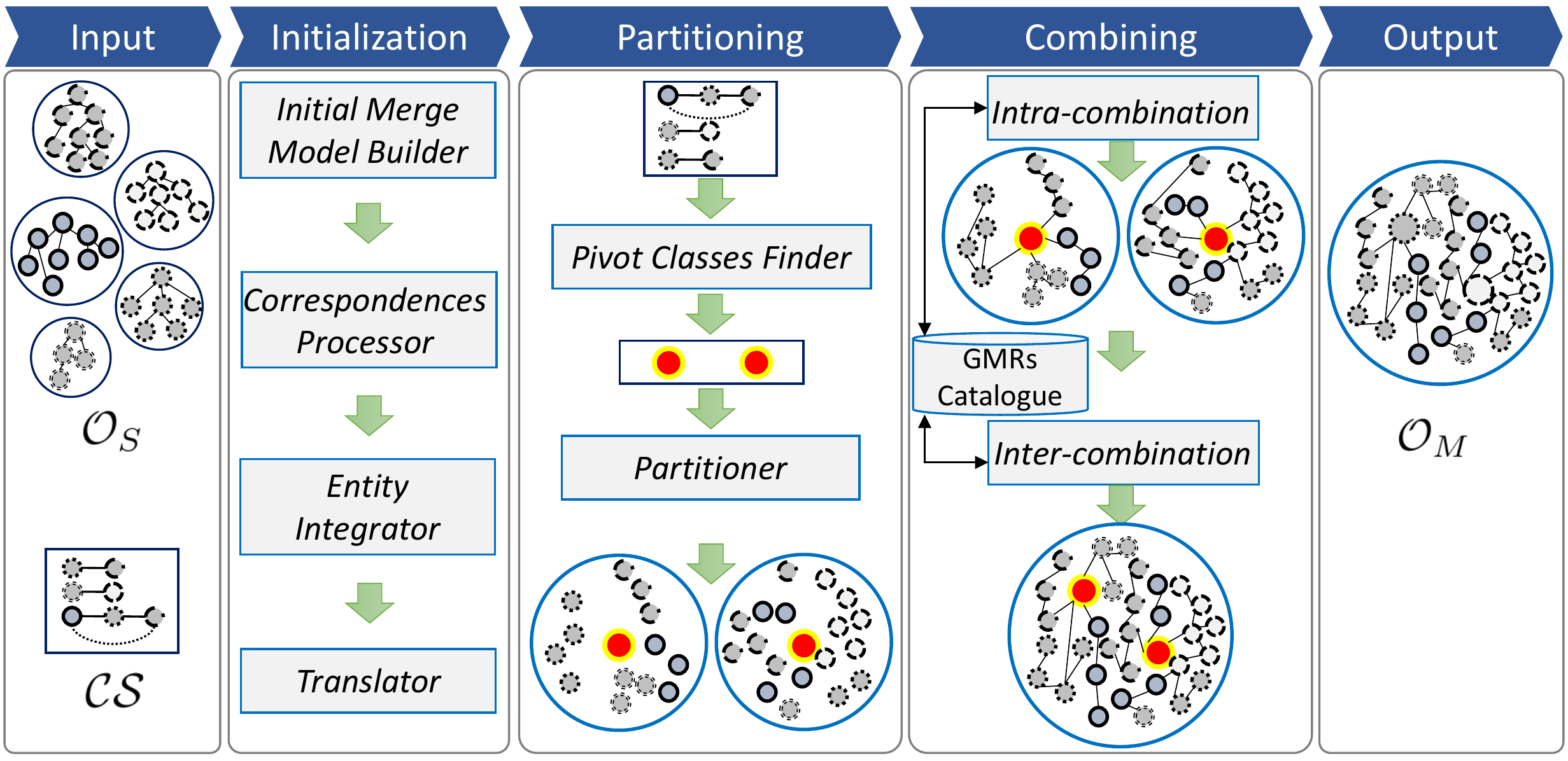}
	\centering
	\caption{The $\mathcal{C}$\textit{o}$\mathcal{M}$\textit{erger} workflow.}
	\label{fig:arc}
	\vspace{-5mm}
\end{figure}

Fig.~\ref{fig:arc} provides an overview of $\mathcal{C}$\textit{o}$\mathcal{M}$\textit{erger}. The input is a set of source ontologies with their correspondence sets, extracted from given mappings, and the merged ontology is the output. In the \textit{Initialization} phase, the source ontologies and the corresponding sets are processed to construct an initial merge model (see Sec.~\ref{sec:Initialization}). In the \textit{Partitioning} phase, the initial merge model, constructed upon the $n$ source ontologies, is divided into $k$ blocks based on structural similarity (see Sec.~\ref{sec:Divide}). Then, in the \textit{Combining} phase, the created blocks are individually refined and finally combined into the merged ontology (see Sec.~\ref{sec:combining}). In the following, we describe preliminaries and each phase in detail.

\subsection{Preliminaries}
An ontology is a formal, explicit description of a given domain. It contains a set of entities $\mathcal{E}$ including classes $C$, properties $P$, and instances $I$. Properties can belong to either taxonomic or non-taxonomic relations. The signature of an ontology constituting on the entity axioms is indicated by $Sig(\mathcal{O})$. 
Since existing mappings are usually generated for pairs of ontologies, we maintain a model $\mathds{M}$ (see Def.~\ref{def:M}), which keeps the information across a group of correspondences over multiple ontologies.
\begin{definition}\label{def:M}
	A model of mappings $\mathds{M}$ over multiple ontologies is built based on the corresponding sets $\mathcal{CS} = \{cs_1,...,cs_t\}$.
	Each element of $\mathds{M}$ holds a set of related corresponding entities between the source ontologies, i.e. $\mathds{M}_i=\{e_1^{\mathcal{O}_j},...,e_d^{\mathcal{O}_h}\}, d \geq 2$, $\mathds{M}_i \in \mathds{M}$, $\mathcal{O}_j$ and $\mathcal{O}_h \in \mathcal{O}_S$.
\end{definition}

We use $cs_j^C$ to denote corresponding classes and $cs_j^P$ for corresponding properties.
For now, we only consider the TBox of ontologies and leave ABox assertions as future work. We use $\equiv$ to represent that two entities are corresponding 
to each other. Suppose the underlying mappings found  $e_1^{\mathcal{O}_1} \equiv e_2^{\mathcal{O}_2}$ and $e_1^{\mathcal{O}_1} \equiv e_3^{\mathcal{O}_3}$. We combine this information into one corresponding set $cs=\{e_1^{\mathcal{O}_1},e_2^{\mathcal{O}_2},e_3^{\mathcal{O}_3}\}$ containing all three entities.
We also denote the cardinality of each corresponding set by $Card(cs)$; in the previous example that is $Card(cs)=3$.
In this context, ontology merging~\cite{vanilla} is then defined as follows:
\begin{definition}
	Ontology merging is the process of creating a new merged ontology $\mathcal{O}_M$ from a set of source ontologies $\mathcal{O}_S=\{\mathcal{O}_1,...,\mathcal{O}_n\}$ given a set of corresponding sets $\mathcal{CS}$ extracted from their mappings $\mathcal{M}$, fulfilling a certain set of requirements.
	\vspace{-2mm}
\end{definition}
The merged ontology is expected to fulfil a set of requirements and criteria assuring its quality.

\subsection{Initialization Phase} \label{sec:Initialization} 
This phase takes as input a set of source ontologies with their corresponding sets and provides an initial merge model $\mathcal{I}_M$. 
Our initialization phase is partially similar to the preliminary process in~\cite{atom} with an extension for multiple ontologies. This phase includes the following tasks: 
\begin{enumerate}
	\item \textbf{Initial merge model builder}: We build an initial empty merge model $\mathcal{I}_M$ and parse the source ontologies to load corresponding and non-corresponding entities into $\mathcal{I}_M$.

	\item \textbf{Correspondences processor}: 
	In this step, the corresponding sets $\mathcal{CS}$ from the given mappings are processed to build the model of mappings $\mathds{M}$ over multiple ontologies.
	If several entities from multiple source ontologies correspond to each other, one joined entry for all 
	is created in this model.

	\item \textbf{Entity integrator}: For each entry of $\mathds{M}$, a new integrated entity in $\mathcal{I}_M$ is created. This means the corresponding entities are combined into a new integrated entity, replacing the original entities in $\mathcal{I}_M$.
	If the original entities within a single set of correspondences have different labels, the newly generated integrated entity will have multiple labels. 
	
	\item \textbf{Translator}: To construct the initial relations between the entities in $\mathcal{I}_M$, we process axioms in $\mathcal{I}_M$. 
	If an axiom's entities have a corresponding entity in $\mathds{M}$, the axiom is translated with the generated integrated entity, i.e., the integrated entity will be replaced with the original entity in each axiom. 
	
\end{enumerate}

The initial merge model $\mathcal{I}_M$ can be used to derive a merge result in a straightforward manner. $\mathcal{C}$\textit{o}$\mathcal{M}$\textit{erger} differs from using $\mathcal{I}_M$ alone, by its focus on applying a set of local and global refinements, including, e.g., structural preservation, acyclicity, or constraint and entailments satisfaction to achieve a quality-assured merged ontology.

\subsection{Partitioning Phase} \label{sec:Divide} 
To partition the source ontologies, we use a set of \textit{pivot} classes $\mathcal{P}$. This is inspired by the work in~\cite{deelers2007enhancing}, where a set of predetermined points has been successfully used in the partitioning method. The partitioning process generates ontologies' blocks, with the following definition:
\begin{definition}
	A block $\mathcal{L}$ is a non-empty subset (or whole) of one (or more) source ontologies $Sig(\mathcal{L}) \subseteq \bigcup\limits_{i=1}^{n} Sig(\mathcal{O}_i), \mathcal{O}_i \in \mathcal{O}_S$ such that every entity from $\mathcal{O}_S$ is in exactly one of these subsets, and the blocks are disjoint.
\end{definition}

The number of blocks is denoted by $k$ and $\mathcal{CL} =\{\mathcal{L}_1, ...,\mathcal{L}_k\}$ is the set of all blocks. In this regard, the ontology merging task $Merge(\mathcal{O}_1, ..., \mathcal{O}_n)$ is decomposed into a block merging task $Merge (\mathcal{L}_1, ..., \mathcal{L}_k)$, where $k<<n$. Thus, the \textit{Partitioner component} in Fig.~\ref{fig:arc} accelerates the merge process by dividing the entities of the initial merge model $\mathcal{I}_M$ into $k$ blocks.
The objective of this partitioning is to maximize intra-block similarity (cohesion) and minimize inter-block similarity (coupling).
This indicates that entities within one block are close to each other in terms of structure\footnote{For each block, a sub-ontology will be created in the intra-combination phase \mbox{(Sec.~\ref{sec:combining}). Thus, intra and inter-similarity can be measured on the sub-ontologies' level.}}, while entities of different blocks are distant from each other. 
We design our partitioning objective function according to this general goal. In software engineering, the notions of cohesion and coupling have been associated with different aspects of software quality (see~\cite{paixao2017empirical} for a survey).
The cohesion represents the high relevance of elements within the same module, while the coupling denotes little relevance of elements across different modules.
Similar to software module measures, ontology module measures are designed to quantify ontology modules' properties.
In the following, we discuss our approach to find pivot classes and our divide method.

\textbf{(i) Finding  pivot classes $\mathcal{P}$.} 
Our method is based on measuring a value for the classes to find $\mathcal{P}$. 
Classes with high $Card(c_t)$ values ($c_t \in \mathcal{CS}$) show high overlap within $\mathcal{O}_S$. Putting them in one block can increase intra-block similarity and achieve our 
partitioning's objective. However, contemplating only this metric tends to choose isolated classes as the pivot classes. To overcome this drawback, the number of connections of each class is taken into account, too.
Thus, the largest sets of corresponding classes in $\mathcal{CS}$ that also have a high number of connections are very promising to be considered as $\mathcal{P}$.
We calculate a \textit{reputation} degree of each class based on the connectivity degree $Conn(c_t)$ and the cardinality of corresponding classes $Card(c_t)$ as given in Eq.~\ref{eq:goodness}. 
Thus, $\mathcal{P}$ is achieved by a sorted list of $\mathcal{CS}$'s elements based on their reputation degrees.
\begin{equation} \label{eq:goodness}
reputation (c_t) = 
Conn(c_t) \times Card(c_t)
\vspace{-2mm}
\end{equation}
The connectivity degree of a class is indicated by the number of associated taxonomic (subClassOf) and non-taxonomic (semantic) relations for the class, $taxo\_rel(c_t)$ and $non\_taxo\_rel(c_t)$, respectively. One can assign different weights for taxonomic or non-taxonomic relations based on the source ontologies' nature. 
Thus, we calculate the connectivity degree of a class $Conn(c_t)$ with user pre-determined weights ($w_t$ and $w_{nt}$) on taxonomy and non-taxonomic relations, as given by Eq.~\ref{eq:connectivity}.
\begin{equation} \label{eq:connectivity}
Conn(c_t) =  w_t \times |taxo\_rel(c_t)| + w_{nt} \times |non\_taxo\_rel (c_t)|
\end{equation}

\textbf{(ii) Partitioner: a structure driven strategy.} This step divides all classes from $\mathcal{I}_M$ into a set of blocks $\mathcal{CL} =\{\mathcal{L}_1, ...,\mathcal{L}_k\}$. For this purpose, we follow a structural-based similarity, in which classes close in the hierarchy are considered similar and should be placed in the same block. Thus, once a class is assigned to a block, all its adjacent classes (on the hierarchy levels of the respective source ontology) consequently will be added. In this regard, the first block $\mathcal{L}_1$ is created by the first $\mathcal{P}$'s element, which has the highest reputation degree. For each corresponding class 
$cs^c_j \in \mathcal{P}$, where $cs^c_j=\{c_1^{\mathcal{O}_i}, ..., c_d^{\mathcal{O}_h}\}$, all classes of $cs^c_j$, i.e., $c_1$ until $c_d$ with all their adjacent classes on their respective ontologies are added to the block. Then, the next element of $\mathcal{P}$ is selected to create a new block, if at least one of its classes has not been assigned to the previous blocks. This process is continued until all elements of $\mathcal{P}$ are processed.
Following this process, the overall number of blocks is automatically determined based on the ratio of the number of $\mathcal{P}$'s elements and amount of overlap 
between $\mathcal{P}$'s elements.

The partitioning process is restricted by two assumptions: (1)~if the taxonomy relation of $\mathcal{P}$'s element is null, no block will be created for it. This prevents the creation of very small blocks
; (2)~if some unconnected classes are left (not belonging to any block), they will not be added to any, since they do not require any refinement in the block. However, the unconnected classes will be added directly to $\mathcal{O}_M$. Overall, our proposed strategy has two advantages: First, it has low computational complexity since it does not need to run a similarity membership function and scales well into a large number of ontologies with many classes. Second, it utilizes the structural similarity between classes by considering the adjacent relationship between classes. Thus, it increases the intra-block similarity of blocks (in terms of hierarchical structure) and significantly reduces the inter-block similarity.

\subsection{Combining Phase} \label{sec:combining}
In this phase, the created blocks are combined to generate the final merged ontology.
To achieve that, we split the combining process into two steps: 

\textbf{(i) Intra-combination: independent merge.} 
In this step, all blocks are processed to be merged and refined. 
Merging the smaller number of blocks reduces the memory allocations compared to merging all source ontologies. This results in a significant reduction of the search space and thus improves efficiency. To further improve performance, the block merging may be performed in parallel. 
Intra-combination parallelization enables the parallel execution of independently executable merges to utilize multiple processors for faster processing.
Thus, the entities inside the blocks are combined to create local sub-ontologies. This step is required to assign the properties for each created block.

In the previous subsection, all the classes have been divided into disjoint blocks. However, these blocks cannot be directly used because the property axioms, which connect these classes, are missing.
Thus, we need to add the properties of the classes to their respective blocks and construct the relationships between classes. We retrieve all axioms from $\mathcal{I}_M$, which already contains the translated corresponding properties. Thus, each class is augmented by the original or translated properties axioms, including all taxonomy and non-taxonomy relations. So, the taxonomy relations between classes as well as non-taxonomic relations are built for each block. A very simple and yet effective approach is to assign each axiom to a block in which all its entities are contained. To keep the blocks disjoint, each axiom should belong to one and only one block. If the classes of each axiom are distributed across multiple blocks, they are not added to any block and are marked as \textit{distributed axioms} ($dist_{axiom}$). 
Their inclusion is  delayed until the next step.

After that, a local refinement process takes place for each sub-ontology. Through our tool, in addition to reusing the final merged ontology, users access the $k$ created local sub-ontologies separately. 
The usage of using sub-ontologies rather than source ontologies has the advantage that the created sub-ontologies present a richer domain knowledge (w.r.t. the knowledge provided by the source ontologies) as they include all similar entities. An additional advantage is that maintaining the source ontologies while keeping the existing mapping between them requires much more effort than keeping the $k$ local sub-ontologies (for which the existing mappings are gathered in one place with limited numbers of mappings between them). So, when local refinements are applied to them, their quality is higher.

\textbf{(ii) Inter-combination: dependent merge.} 
In the \textit{inter-combination} step, the global merged ontology $\mathcal{O}_M$ is constructed based on the $k$ created local sub-ontologies. 
For this to be achieved, we follow a sequential merge processing in this step based on the inter-block relatedness degree, which represents how much two blocks differ from each other. Thus, we calculate the number of shared distributed axioms $dist_{axiom}$ between two blocks $\mathcal{L}_i $ and $\mathcal{L}_j$ of $\mathcal{L}$ to indicate the inter-block relatedness, as shown in Eq.~\ref{eq:interSim}. 
\begin{equation} \label{eq:interSim}
inter\_rel(\mathcal{L}_i,\mathcal{L}_j) = |dist_{axiom}(\mathcal{L}_i) \cap dist_{axiom}(\mathcal{L}_j)|
\end{equation}

At first, the two blocks $\mathcal{L}_i$ and $\mathcal{L}_j$ with the highest inter-block relatedness value 
are merged into $\mathcal{L}_{ij}$. This includes adding all distributed axioms to them. Then, the next block, which has the highest inter-block relatedness value with the recently merged block $\mathcal{L}_{ij}$, will be merged. After merging blocks, the number of distributed axioms between the recent merged one and the remaining blocks will be updated. This process is supported so that the most similar blocks can be executed earlier, and less similar blocks are processed at later steps. Note that, the approach in~\cite{coma} also matched only similar blocks and delayed the processing of dissimilar blocks. The sequential execution of the merging will be continued until all blocks are processed. 
Note that, two blocks are by nature disjoint and no shared classes exist on both. However, the inter-relatedness for them might not always be zero, since this metric is based on the number of distributed axiom between them.
If the inter-block relatedness between two blocks is zero, they will be entirely disjoined and will not need any merge process. Thus, they will be imported directly to the  $\mathcal{O}_M$. 

A set of global refinements will be applied on the last combined block. Upon that, in the last step, the merged ontology $\mathcal{O}_M$ is built. 
The reason to merge the most similar blocks first is that the most similar blocks have much more distributed axioms. Combining these two blocks in the earlier steps is more efficient when the blocks are small. Note that, in each sequential merge, the intermediate merged blocks will get larger. So, it is more efficient in the number of processes to have less processing when the blocks get more massive. 

To apply local and global refinement within $\mathcal{C}$\textit{o}$\mathcal{M}$\textit{erger}, we utilize a list of General Merge Requirements (GMR)s, which were introduced in our previous work~\cite{samiraGMR}. From these GMRs, users can select a subset to be applied\footnote{For details of applying GMRs, see: http://comerger.uni-jena.de/requirement.jsp} according to their requirements. Fulfilling each GMR making sure that the merged ontology meets the chosen GMRs, if needed by adapting it, contributes to its refinement. This leads to a generic, flexible parameterizable merge method. Moreover, users can easily adjust this framework so that different refinements perform in intra- and inter-combination steps.

\section{Experimental Evaluation} \label{sec:result}
To validate the applicability of the proposed approach, we conducted a series of experiments utilizing different sets of ontologies to analyze quality, runtime, and complexity performance. 
The proposed approach has been implemented in $\mathcal{C}$\textit{o}$\mathcal{M}$\textit{erger}~\cite{samiraTool2020}, a tool that is publicly available on~\url{http://comerger.uni-jena.de/} and distributed\footnote{https://github.com/fusion-jena/CoMerger} under an open-source license along with the merged ontologies. 
In the following, we present the used datasets, describe experimental environments, and report on the results.

\textbf{3.1 $\ \ \ $ Datasets.}
To evaluate the general applicability of our approach, we have aimed to use a wide variety of datasets both in terms of subject and in terms of size in our experiments. 
We have selected sets of ontologies\footnote{https://github.com/fusion-jena/CoMerger/blob/master/MergingDataset/datasetInfo.md} 
from the conference ($d_1$-$d_6$), anatomy ($d_7$) and large biomedical ($d_8$) tracks of the OAEI benchmark\footnote{http://oaei.ontologymatching.org/2019/} along with ontologies from BioPortal\footnote{https://bioportal.bioontology.org/; accessed at 01.10.2019} in the domains of biomedicine ($d_9$), and health ($d_{10}$), the union of both ($d_{11}$) as well as combination of several subdomains ($d_{12}$). Our dataset includes a variety of ontologies with different axioms' size ($134 \leq |Sig(\mathcal{O}_S)| \leq30364$) and numbers of source ontologies  ($2 \leq n \leq 55 $). We conducted our tests with two different types of correspondences (mappings): 
(i)~a perfect mapping  $\mathcal{M}$ from the OAEI benchmark and BioPortal's mappings, (ii)~an imperfect mapping $\mathcal{M'}$ which is produced by an ontology matching system~\cite{seecont}. While the first shows the general potential of the approach, the latter shows its applicability in a realistic setting where typically no perfect mapping is available.

\textbf{3.2 $\ \ \ $ Test Setting.} \label{sec:setting}
All the experiments were carried out on Intel core i7 with 12 GB internal memory on Windows 7 with Java compiler 1.8. The values of $w_t$ and $w_{nt}$ were empirically determined to 0.75 and 0.5, respectively, but we make no claim that these are optimal values. We evaluated our approach under different conditions $V_1$-$V_{12}$ (see Table~\ref{table:setting}). This includes using the perfect mapping $\mathcal{M}$ vs. an imperfect mapping $\mathcal{M'}$ and applying the refinement process on only the local level, only the global level and on both levels. Thus, we generated six versions ($V_1$-$V_6$) of merged ontology using the n-ary method. 
We ran for each dataset two types of binary merges (\textit{balanced} and \textit{ladder}~\cite{batini1986comparative}) under the name $V_7$-$V_9$ and $V_{10}$-$V_{12}$, respectively. For them, we consider the imperfect mapping\footnote{At each merge process, the mapping for the created intermediate merged result and one of $\mathcal{O}_S$ is generated on the fly with the ontology matching tool.}. 
For the \textit{ladder} binary merge, a new ontology is integrated with an existing intermediate result at each step. For the \textit{balanced} binary strategy, the ontologies are divided into pairs at the start and are integrated in a symmetric fashion.

\textbf{Refinement Setting.}
To apply local and global refinements, we select a subset of GMRs (\textit{R1}-\textit{R3}, \textit{R7}, \textit{R15}, \textit{R16}, \textit{R19}) from~\cite{samiraGMR}. \textit{R1}, \textit{R2}, \textit{R3}, and \textit{R7} are related to class, property, instance, and structure preservation, respectively. \textit{R15} restricts properties without multiple domains or ranges, so-called oneness characteristics. \textit{R16} is relevant to class acyclicity and \textit{R19} expresses the degree of connectivity in $\mathcal{O}_M$. We use these criteria to observe how well the merged ontologies are structured. The remaining GMRs do not have special effect on our datasets, so we do not present them here.

\begin{table}[bt!]
	\caption{The settings for generating twelve variants of the merged ontologies.}
	\centering
	\resizebox{7cm}{!} {
		\begin{tabular}{| c ? c | c | c | c | c | c ? c | c | c ? c | c | c |}
			\hline
			\rowcolor{gray!15}
			\multirow{2}{*}{} & \multicolumn{6}{c?}{N-ary} & \multicolumn{3}{c?}{Balanced} & \multicolumn{3}{c|}{Ladder} \\ \cline{2-13}\omit \vrule height.4pt\textcolor{gray!15}{\leaders\vrule\hfil}\vrule \cr 
			\rowcolor{gray!15}
			& $V_1$ &$V_2$ &$V_3$ & $V_4$ & $V_5$ &$V_6$ & $V_7$& $V_8$ & $V_9$ & $V_{10}$ & $V_{11}$ & $V_{12}$\\ [0.4ex]
			\Xhline{3\arrayrulewidth} 
			\cellcolor{gray!15}Mapping type & $\mathcal{M}$ & $\mathcal{M}$ & $\mathcal{M}$ & $\mathcal{M'}$ & $\mathcal{M'}$& $\mathcal{M'}$ &$\mathcal{M'}$&$\mathcal{M'}$ & $\mathcal{M'}$ & $\mathcal{M'}$ & $\mathcal{M'}$ & $\mathcal{M'}$\\		\hline
			\cellcolor{gray!15}Global refinement & $\checked$ &$\checked$& $\times$ & $\checked$&$\checked$ &$\times$ & $\checked$&$\checked$ &$\times$ &$\checked$&$\checked$&$\times$\\		\hline
			\cellcolor{gray!15}Local refinement & $\checked$ & $\times$ & $\times$ &$\checked$ &$\times$ &$\times$ &$\checked$&$\times$ &$\times$&$\checked$&$\times$&$\times$\\		\hline
	\end{tabular}}
	\label{table:setting}
	\vspace{-4mm}
\end{table}

\begin{figure}[tb!]
	\includegraphics[width=12cm,keepaspectratio]{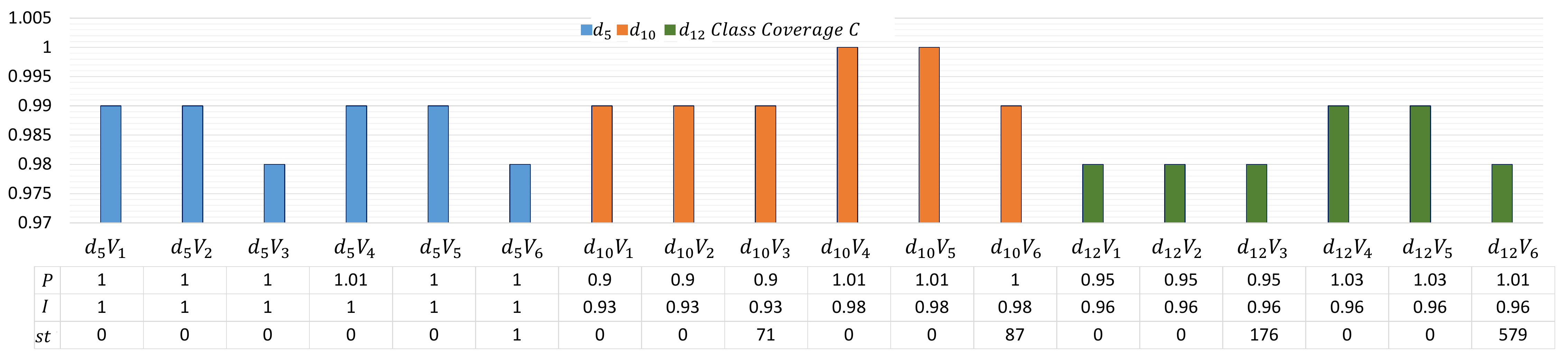}
	\centering
	\caption{Class $C$, property $P$ and instance $I$ coverage with the number of unpreserved structure $str$ of six versions of n-ary merge for 3 sample datasets.}
	\label{fig:coverage}
	\includegraphics[width=12cm,keepaspectratio]{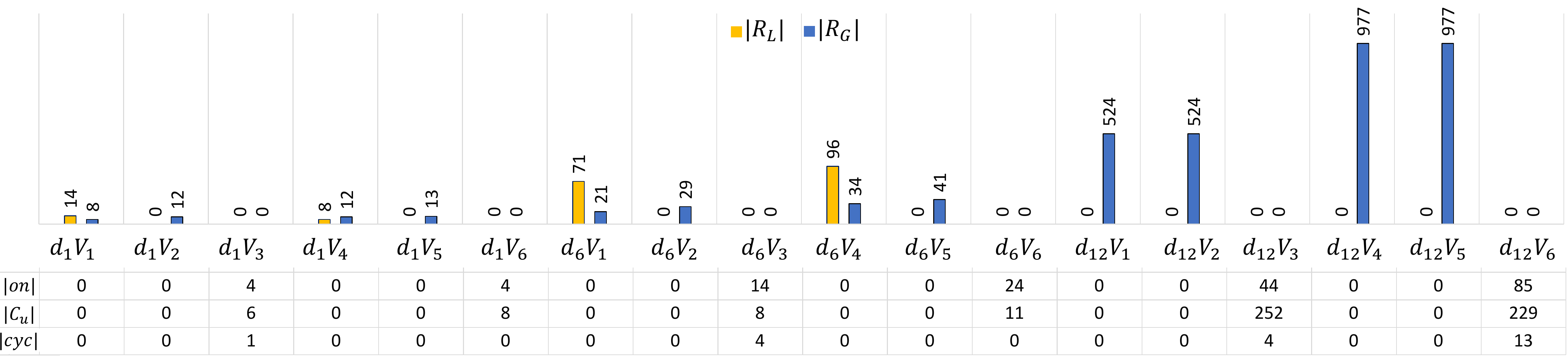}
	\centering
	\caption{Number of local $|R_L|$ and global $|R_G|$ refinements, oneness $|on|$, unconnected classes $|C_u|$ and cycle $|cyc|$ of six versions of n-ary merge.}
	\label{fig:ref}
	\vspace{-4mm}
\end{figure}

\textbf{3.3 $\ \ $ Experimental Results.} \label{sec:ExperimentalRes}
In the first test, we observe the characteristics of the n-ary merged ontologies. 
In the second test, we analyze the constructed logic of the merged ontology by answering a group of Competency Questions (CQs). 
Comparing binary merge and n-ary methods is demonstrated in the third test. 
For an inconsistency test (related to the model's entailment), we refer the readers to our previous work~\cite{SamiraSLogic}. Due to limited space, showing the results of all datasets with all twelve different variations is not possible in this paper. 
Thus, in Fig.~\ref{fig:coverage}, Fig.\ref{fig:ref} and Table~\ref{table:binHol}, we show the results on some of the datasets only. Our discussion takes the full set of experiments into account, though. The corresponding results are available in our repository\footnote{https://github.com/fusion-jena/CoMerger/blob/master/MergingDataset/result.md} and in the appendix.

\textbf{First Test: Characteristics of the N-Ary Merged Result.} To evaluate the characteristics of the created $\mathcal{O}_M$, we use three evaluation criteria categories: 
\begin{itemize}
	\item  \textit{Integrity}: in~\cite{duch2010}, the integrity of a merged ontology is defined as its compactness, completeness, and redundancy. \textit{Compactness} represents the size of the merge result\footnote{It is not presented in this paper, but is available in our repository.}. 
	\textit{Coverage} or namely \textit{completeness} 
	is the percentage of entities present in the $\mathcal{O}_S$ that are included in $\mathcal{O}_M$. 
	This includes classes $C$, properties $P$, instances $I$, and the structurality $str$ coverage. The latter one refers to preserving the structure of the merged ontology w.r.t. source ontologies. These metrics are related to the evaluation of \textit{R1}-\textit{R3}, and \textit{R7} from~\cite{samiraGMR}. \textit{Redundancy} 
	checks whether redundant entities appear in $\mathcal{O}_M$. Since we found no redundant entities in any of the created versions of the merged ontologies, 
	we do not include this metric in results. 
	
	\item \textit{Evaluation of applying the GMRs}: we evaluate to which extent the refinements play a role. For \textit{R15} (oneness), we count the number of properties that have multiple domains or ranges and present it as $|on|$. For \textit{R19} (connectivity), we consider only those unconnected classes in the $\mathcal{O}_M$ which were connected in the $\mathcal{O}_S$ given by $|C_u|$.		
	For \textit{R16} (acyclicity), we calculate how many cycles in the class hierarchy in the $\mathcal{O}_M$ exist ($|cyc|$). 
	
	\item \textit{Merge process characteristic}: we address the characteristic of the merge process by measuring the number of created blocks $k$, percentage of distributed is-a ($ds\%$) and the translated ($tr\%$) axioms on the total axioms, number of local $|R_L|$ and global refinement $|R_G|$ actions in intra- and inter-combinations.
\end{itemize}

In Fig.~\ref{fig:coverage}, we show the degree of information preservation on the six versions of our n-ary merge method. 
The percentage of class coverage $C\%$ is shown on the chart, while the percentage of the property $P\%$, instance $I\%$ coverage, and the absolute number of unpreserved structure $|str|$ are drawn in the table view under each chart. 
In Fig.~\ref{fig:ref}, we show the number of local $|R_L|$ and global $|R_G|$ refinements actions. 
In this figure, we also present the statistics about the evaluation of selected GMRs with $|on|$, $|C_u|$, and $|cyc|$. 

To analyze the result of these two figures, we examine the result of different versions. By investigating the effect of considering no refinements ($V_3$/$V_6$) compared to applying refinements either locally or globally ($V_1$,$V_2$,$V_4$,$V_5$), we can conclude that applying refinement in 76 out of 120 cases leads to considerable improvements in class coverage, structure preservation, oneness, unconnected classes, and acyclicity, whereas in 40 cases, the same results are achieved. For example, if no refinement is applied, 8 classes became unconnected in $d_{1}$, or 24 properties with multiple domains and ranges in $d_6$ are generated, or 13 cycles exist in $d_{12}$, or 71 unpreserved structures happen in $d_{10}$. In comparing usage of perfect or imperfect mappings, we observe that out of 12 datasets, a perfect mapping causes 7 fewer cases of unconnected entities (e.g., in $d_1$ and $d_6$); 3 fewer cases of cycles (e.g., in $d_{12}$); 7 fewer cases of properties oneness (e.g, in $d_{12}$); 6 cases preserving better structure (e.g., in $d_{10}$ and $d_{12}$); 3 fewer cases of local refinements (e.g., in $d_6$); 9 fewer cases of global refinements (e.g., in $d_1$, $d_6$, and $d_{12}$). On the other hand, using imperfect mapping causes 1 fewer case of unconnected entities in $d_{12}$; 6 fewer cases of cycles (e.g., in $d_1$ and $d_6$); 4 fewer cases of local refinements (e.g., in $d_1$); and 2 fewer cases of global refinements. The effect of applying local refinements $V_1/V_4$ rather than no local refinements $V_2/V_5$ cannot be observed in the view of the mentioned criteria. However, applying refinement actions in the local or global level have different computational complexity, since the respective search spaces substantially differ. For instance, finding or repairing a cycle in a small set of classes (local sub-ontologies) is far less expensive than among all classes of the merged ontology.

\begin{figure}[tb!]
	\includegraphics[width=10cm,keepaspectratio]{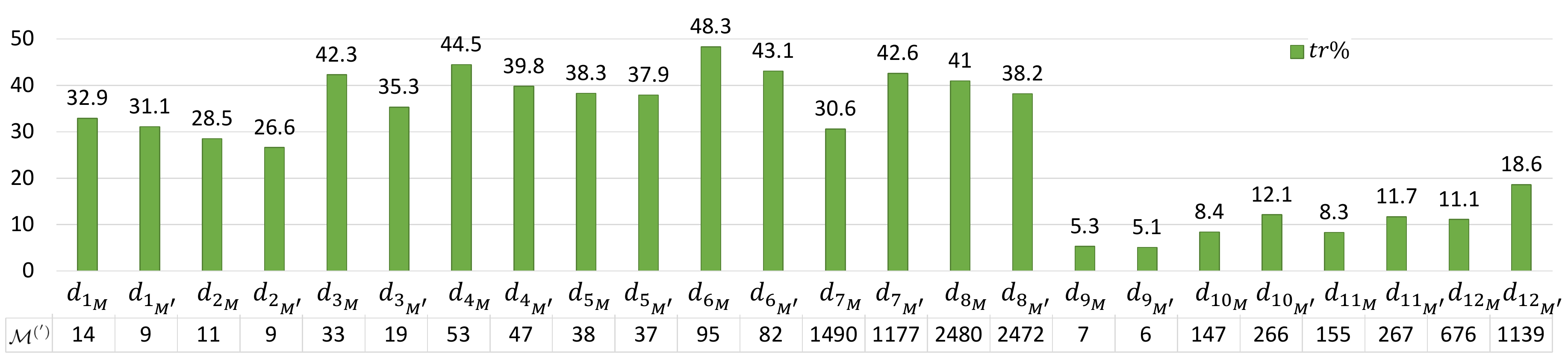}
	\centering
	\caption{\mbox{Comparing translated axioms $tr\%$ with corresponding entities in $\mathcal{M}$ and $\mathcal{M}'$.}}
	\label{fig:translate}
	\includegraphics[width=10cm,keepaspectratio]{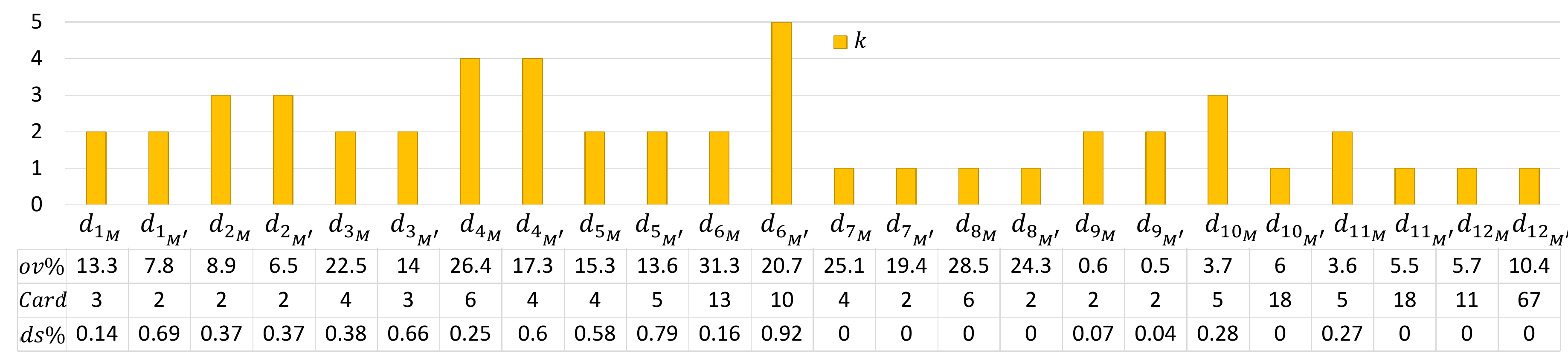}
	\centering
	\caption{Number of blocks $k$ vs. class overlap $ov\%$, max cardinality $Card$, and distributed axioms $ds\%$.}
	\label{fig:kEffect}
	\vspace{-8mm}
\end{figure}

In Fig.~\ref{fig:translate}, we show the percentage of translated axioms for all datasets along with the number of corresponding entities in perfect $\mathcal{M}$ or imperfect $\mathcal{M}'$ mapping. Overall, using perfect or imperfect mappings with different numbers of corresponding entities has a direct effect on the number of translating $tr$ axioms.

In Fig.~\ref{fig:kEffect}, we demonstrate the number of created blocks $k$ for all datasets using perfect $\mathcal{M}$ ($V_1$-$V_3$) or imperfect $\mathcal{M}'$ mappings ($V_4$-$V_6$). The value of $k$ affects the number of distributed axioms. Thus, we also report the percentage of the distributed taxonomic (is-a) axioms on the total axioms $ds\%$. 
These axioms mostly relate to the axioms with \textit{objectUnionOf}, where the union classes are distributed over the blocks.
Determining $k$ in our method mainly depends on the amount of overlap $ov\%$ between $\mathcal{O}_S$'s classes and the cardinality $Card$ value on corresponding classes. The overlap is calculated by the ratio of the numbers of corresponding classes on the total classes. For each dataset, we show the maximum cardinality between the corresponding entities. 
Considering $ov\%$ and $Card$, the values of $k$ in our datasets are reasonable ($1 \leq k \leq 5$), which it shows the feasibility of our approach.

\begin{table}[bt!]
	\caption{Answering CQs on the different versions of merged ontologies.} 
	\centering
	\resizebox{7cm}{!} {
		\begin{tabular}{| c | c | c | c | c | c | c || c |}
			\hline
			\rowcolor{gray!15}
			& &$Semi$\% & & &  && \textcolor{blue}{$Total$} \\ 
			\rowcolor{gray!15}
			\multirow{-2}{*}{id} &\multirow{-2}{*}{$Complete$\%}&-$Complete$ &\multirow{-2}{*}{$Partial$\%}&\multirow{-2}{*}{$Wrong$\% } &\multirow{-2}{*}{$Unknown$\%} &\multirow{-2}{*}{ $Null$\% } &\textcolor{blue}{$Correct$\%} \\
			[0.4ex]
			\Xhline{3\arrayrulewidth}
			\cellcolor{gray!15}$d_1V_1,V_2$  & \multirow{4}{*}{\textbf{20}}              & 6.7                & \textbf{3.3}           &      \multirow{4}{*}{\textbf{3.3}}    & 66.7          & \textbf{0}          & \multirow{4}{*}{\textcolor{blue}{\textbf{30}}}   \\ \hhline{|-~|--~|--~|}
			\cellcolor{gray!15}$d_1V_3$  &               & \multirow{3}{*}{\textbf{10}}                 & \multirow{3}{*}{0}             &          & \textbf{63.3}          & 3.4        &    \\ 
			\hhline{|-~|~|~|~|--~|}
			\cellcolor{gray!15}$d_1V_4,V_5$  &               &                  &              &          & 66.7          & \textbf{0}          &    \\ 
			\hhline{|-~|~|~|~|--~|}
			\cellcolor{gray!15}$d_1V_6$  &               &                  &              &         & \textbf{63.3}          & 3.4        &    \\ \Xhline{3\arrayrulewidth}
			\cellcolor{gray!15}$d_2V_1$-$V_6$  & \textbf{13.3}            & \textbf{6.7}                & \textbf{13.3}          & \textbf{0}           & \textbf{66.7}          & \textbf{0}          & \textcolor{blue}{\textbf{33.3}} \\ 
			\Xhline{3\arrayrulewidth}
			\cellcolor{gray!15}$d_3V_1,V_2$  & 36.7            & 26.7               & \textbf{6.7}           & \multirow{2}{*}{\textbf{0}}           & \multirow{8}{*}{\textbf{26.7}}          & \textbf{3.2}        & \textcolor{blue}{\textbf{70}}   \\ 
			\hhline{|----~|~--|}
			\cellcolor{gray!15}$d_3V_3$  & 33.3            & \textbf{30}                 & \multirow{4}{*}{3.3}           &            &           & 6.7        & \multirow{2}{*}{\textcolor{blue}{66.7}} \\ 
			\hhline{|---~|-~|-~|}
			\cellcolor{gray!15}$d_3V_4,V_5$  & \textbf{40}              & \multirow{5}{*}{23.3}               &            & \multirow{3}{*}{3.3}         &           & 3.4        &  \\ \hhline{|--~|~|~|~|--|} 
			\cellcolor{gray!15}$d_3V_6$  & 36.7            &                &            &         &           & 6.7        & \textcolor{blue}{63.3} \\ \hhline{|--~|~|~|~|--|}
			\cellcolor{gray!15}$d_3V_7,V_8$  & \textbf{40}              &               &            &          &           & 3.4        & \textcolor{blue}{66.7} \\ \hhline{|--~|~|~|~|--|}
			\cellcolor{gray!15}$d_3V_9$  & 36.7            &                &            &          &           & 6.7        & \textcolor{blue}{63.3} \\ \hhline{|-----~|--|}
			\cellcolor{gray!15}$d_3V_{10},V_{11}$ & 26.7            & \multirow{2}{*}{13.3}               & \multirow{2}{*}{0}             & \multirow{2}{*}{10}          &         & 23.3       & \textcolor{blue}{40}   \\ \hhline{|--~|~|~|~|--|}
			\cellcolor{gray!15}$d_3V_{12}$ & 23.3            &                &              &           &           & 26.7       & \textcolor{blue}{36.7} \\
			\Xhline{3\arrayrulewidth}
			\cellcolor{gray!15}$d_4V_1$-$V_3$  & 33.3            & \textbf{20}                 & 10            & \textbf{0}           & 36.7          & \multirow{2}{*}{\textbf{0}}          & \multirow{2}{*}{\textcolor{blue}{\textbf{63.3}}} \\ \hhline{|------~|~|}
			\cellcolor{gray!15}$d_4V_4$-$V_6$  & \textbf{36.7}            & \multirow{2}{*}{13.3}               & \multirow{2}{*}{\textbf{13.3}}          & \multirow{2}{*}{3.3}         & \multirow{3}{*}{\textbf{33.4}}          &           &  \\ \hhline{|--~|~|~|~|--|}
			\cellcolor{gray!15}$d_4V_7$-$V_9$  & 33.3            &                &           &          &          & 3.4        & \textcolor{blue}{60}   \\ \hhline{|-----~|--|}
			\cellcolor{gray!15}$d_4V_{10}$-$V_{12}$ & 23.3            & 10                 & 3.3           & 10          &           & 20         & \textcolor{blue}{36.7} \\
			\Xhline{3\arrayrulewidth}
			\cellcolor{gray!15}$d_5V_1$-$V_3$  & \multirow{3}{*}{\textbf{23.3}}            & 13.4               & \multirow{2}{*}{\textbf{23.3}}          & \multirow{3}{*}{\textbf{0}}           & 40            & \multirow{2}{*}{\textbf{0}}          & \textcolor{blue}{60}   \\ \hhline{|-~|-~|~|-~|-|} 
			\cellcolor{gray!15}$d_5V_4$-$V_9$  &             & \textbf{16.7}               &           &            & \multirow{2}{*}{\textbf{36.7}}          &          & \textcolor{blue}{\textbf{63.3}} \\ \hhline{|-~|--~|~|--|}
			\cellcolor{gray!15}$d_5V_{10}$-$V_{12}$ &             & 13.3               & 16.7          &            &           & 10         & \textcolor{blue}{53.3} \\
			
			\Xhline{3\arrayrulewidth}
			\cellcolor{gray!15}$d_6V_1$-$V_3$  & \multirow{2}{*}{\textbf{40}}              & \multirow{2}{*}{\textbf{23.3}}               & \textbf{10}            & \multirow{2}{*}{\textbf{0}}           & 26.7          & \textbf{0}          & \textcolor{blue}{\textbf{73.3}} \\ \hhline{|-~|~|-~|---|}
			\cellcolor{gray!15}$d_6V_4$-$V_9$  &               &                & 6.7           &            & \multirow{2}{*}{\textbf{23.3}}          & 6.7        & \textcolor{blue}{70}   \\ \hhline{|-----~|--|}
			\cellcolor{gray!15}$d_6V_{10}$-$V_{12}$ & 33.3            & 6.7                & 0             & 6.7         &           & 30         & \textcolor{blue}{40}   \\ \hline
	\end{tabular}}
	\label{table:CQ_noChanges} 
	\vspace{-6mm}%
\end{table}

\textbf{Second Test: Answering Competency Questions (CQs).}
CQs are a list of questions used in the ontology development life cycle, which an ontology should answer. By using Competency Questions tests, we aim to observe which created $\mathcal{O}_M$ can provide 
superior answers to the CQs.
To this end, we used a set of CQs (available in our portal) in the conference domain. Each CQ has been converted manually to a SPARQL query and run against the $\mathcal{O}_S$ and the different versions of the $\mathcal{O}_M$ (our datasets in the conference domain). We compare the CQ-results for each dataset with all possible answers from the $\mathcal{O}_M$ with respect to its $\mathcal{O}_S$ on that dataset. The \textit{complete} answer indicates a full answer.
Among all answers of the $\mathcal{O}_S$, if the number of found answers in $\mathcal{O}_M$ is higher (lower) than the number of not found, we marked it as a \textit{semi-complete} (\textit{partial}) answer. 
An answer is marked as \textit{wrong} if CQ on the $\mathcal{O}_M$ does not return the same answer as the source ontologies. 
If CQ's entities exist in the ontology, but no further knowledge exists about them, we mark them by a \textit{null} answer. If the ontology does not have any knowledge about the CQ, we indicate this by an \textit{unknown} answer. The results are presented in~Table~\ref{table:CQ_noChanges}, where the values are shown as the percentage of the total number of CQs. Values in boldface show the best result (highest values in \textit{complete}, \textit{semi-complete}, \textit{partial}, and \textit{total-complete} answers and lowest values in \textit{wrong}, \textit{unknown}, and \textit{null} answers) in each dataset. The last column shows a sum value on the \textit{complete}, \textit{semi-complete}, and \textit{partial}
answers given by the \textit{total correct} answer.

Overall the result in Table~\ref{table:CQ_noChanges} indicates that 
applying local or global refinements in some cases can provide more \textit{complete} answers (cf. $d_3V_1,V_2$ and $d_3V_4V_5$), more \textit{partial} answers (cf. $d_1V_1,V_2$ and $d_3V_1,V_2$), and less \textit{null} answers (cf. in $d_1V_1,V_2$, $d_1V_4,V_5$, and $d_3V_1,V_2$). 
Using perfect mappings causes more \textit{semi-complete} answers in $d_3$ and $d_4$, more \textit{partial} answers in $d_1$ and $d_3$, less wrong answers in $d_3$ and $d_4$, and less \textit{null} answers in $d_3$ and $d_6$. On the other hand, imperfect mappings have more \textit{complete} answers in $d_3$ and $d_4$, more \textit{semi-complete} answers in $d_5$, more \textit{partial} answers in $d_4$, less unknown answers in $d_4$-$d_6$, less \textit{null} answers in $d_4$, and less \textit{unknown} answers in $d_6$. One of the reasons why the perfect mappings in some cases are worse than imperfect mapping is, the entities inside the ontology in OAEI mapping correspond to each other. This self-mapping causes the structure of the $\mathcal{O}_M$ to be mixed up and makes it unable to provide correct CQs' answering. 

Comparing binary-balanced ($V_7$-$V_9$) and binary-ladder ($V_{10}$-$V_{12}$) shows that in all cases different results are generated. Indeed, the difference between two versions of the binary merges is mostly related to the order of selecting and merging them, which affects the final merged output. Consequently, the answers to the CQs are different. 
Comparing the n-ary ($V_4$-$V_6$) and binary ($V_7$-$V_{12}$) strategies reveals that the n-ary merge can achieve the same quality result as binary methods, and even better results in $d_4$ in terms of achieving more \textit{complete} answers and less \textit{null} answers rather than binary approaches.
Summing up, the ontologies in the conference domain are small in size and lack complete knowledge modeling. Because of this, 100\% complete answers are hard to  achieve. However, our merged ontologies can provide up to $73.3\%$ totally correct answers, which shows the applicability of our method. 

\begin{table}[tbp]
	\captionof{table}{Comparing n-ary (N), balanced (B) and ladder (L) merge strategies with the number of corresponding entities $|Cor|$, translated axioms $|tr|$, global refinements $|R_G|$, merge processes $|Mer.|$}
	\centering
	\resizebox{\textwidth}{!} {
		\begin{tabular}{|c ? c|c|c ? c|c|c ? c|c|c ? c|c|c ? c|c|c|}
			\hline
			\rowcolor{gray!15}
			& \multicolumn{3}{c?}{$d_4$}& \multicolumn{3}{c?}{$d_6$}& \multicolumn{3}{c?}{$d_{10}$}& \multicolumn{3}{c?}{$d_{11}$}& \multicolumn{3}{c|}{$d_{12}$} \\ \hhline{|~|---------------|}
			\rowcolor{gray!15}
			& N & B & L & N & B & L & N & B & L & N & B & L & N & B & L   \\ \Xhline{3\arrayrulewidth} 
			\cellcolor{gray!15}$|Cor|$ & 47    & 65       & 64     & 82    & 127      & 128    & 266   & 369      & 351    & 267   & 388      & 386    & 1139  & 2186     & 2159   \\ \hline
			\cellcolor{gray!15}$|tr|$& \ 790 \   &  \ 1270   \    &  \ 1339  \   &  \ 1310  \  & 2791 \      &  \ 3462   \  &  \ 6949 \   & \  15816   \   & \  31420  \  & \  6960 \   &  \ 16060 \     & \  35480 \   &  \ 30035 \  &  \ 66344  \    & \  154002 \  \\ \hline
			\cellcolor{gray!15}$|R_G|$& 21    & 27       & 23     & 41    & 48       & 51     & 143   & 166      & 191    & 143   & 177      & 196    & 977   & 1533     & 1560   \\ \hline
			\cellcolor{gray!15}$|Mer.|$ & 1&\multicolumn{2}{c?}{3}&1&\multicolumn{2}{c?}{6}&1&\multicolumn{2}{c?}{16}&1&\multicolumn{2}{c?}{18}&1&\multicolumn{2}{c|}{54} \\   \hline
	\end{tabular}}
	\label{table:binHol}
	\vspace{-4mm}%
	\includegraphics[width=10cm,keepaspectratio]{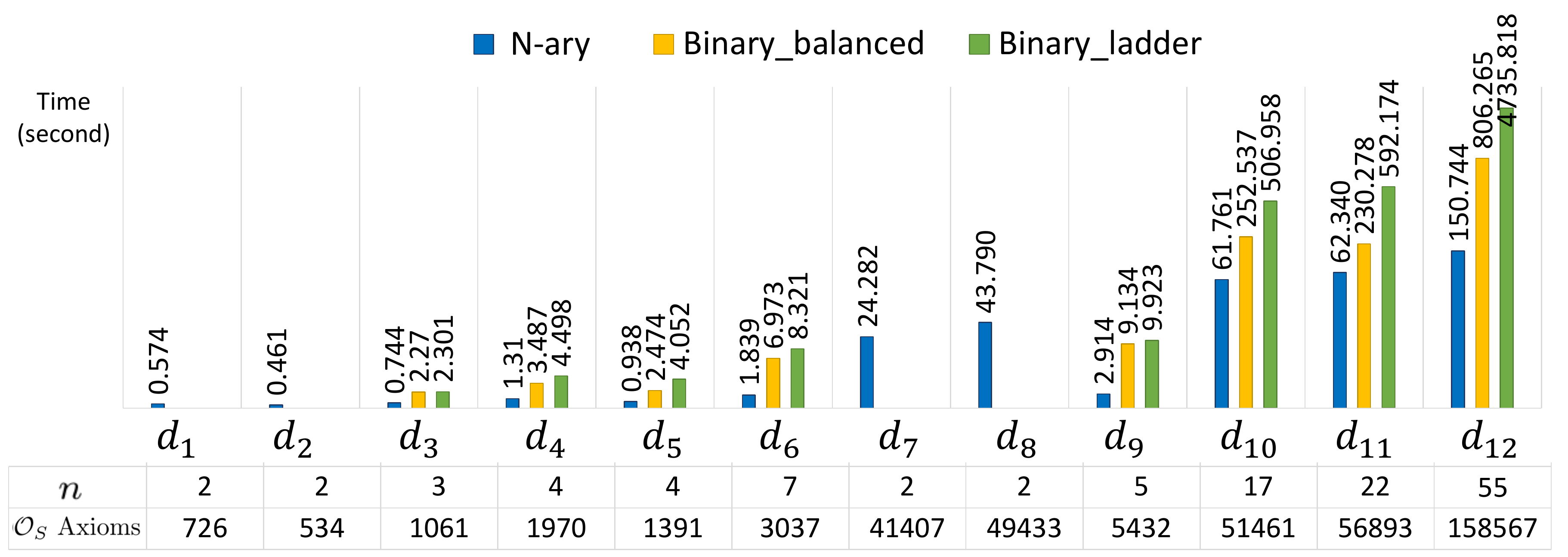}
	\centering
	\captionof{figure}{Runtime performance: numbers of source ontologies and axioms versus the required time for the merge process in second. Binary merges run on datasets with $n>2$. }
	\label{fig:time}
	\vspace{-8mm}%
\end{table}

\textbf{Third Test: Binary vs. N-ary.} 
While the CQ test shows that the result quality of the n-ary approach can compete with the binary approaches, in the third test, we compare performances metrics. 
We conduct an experimental test by a series of binary merges on the eight datasets that have more than two source ontologies. 
We examine the runtime performance and required operations. 
Table~\ref{table:binHol} shows the difference of operation complexities between these three strategies, where due to better legibility, only some datasets in comparing of n-ary ($V_5$), balanced ($V_8$), and ladder ($V_{11}$) are shown. However, in interpreting the results, we make conclusion on the all eight datasets with versions $V_4$-$V_{12}$.  
The number of total corresponding entities $|Cor|$ during the whole process of the merge is quite different.
In each test of a binary merge, only the correspondences between two entities can be integrated into a new entity. However, in n-ary merge, the corresponding entities from multiple source ontologies can be integrated simultaneously into the new entity. For this reason, the number of corresponding entities in the binary merge in 7 out of 8 datasets is much higher than the n-ary approach. Consequently, the required amount of combining them into new entities and translating their axioms $|tr|$ is high in all tested datasets. For instance, in $d_6$, the number of translated axioms in the n-ary method is 1310, while in the binary-ladder strategy it is 3462. Therefore, the n-ary approach has great speed-up.

We compare the number of required refinement actions $|R_G|$ in Table~\ref{table:binHol}. In 7 out of 8 datasets, the n-ary approach requires fewer refinement actions compared to binary merges. For instance, in $d_{12}$, the n-ary method runs 977 actions, while in the  binary-balanced is 1533. The same conclusion can be derived from comparing the required local refinements. We also present the number of merge processes by $|Mer.|$ in the table. 
While the n-ary approach only uses one iteration for all tests, ladder and balanced methods require $n-1$ merge process, e.g., in $d_{12}$, 54 times the whole process of the merge should be run.

We demonstrate the method's \textit{scalability} by illustrating the performance test results. Here, the runtime performance is evaluated based on the number of ontologies versus the required time for the merge process in n-ary ($V_5$), balanced ($V_8$), and ladder ($V_{11}$). Fig.~\ref{fig:time} shows the total runtime in seconds for the merge processes. We ran each test 10 times and presented the average values. 
The processing time of the binary merge does not include the time for creating the respective mappings. In this test, we present the number of $\mathcal{O}_S$ and their axioms, to emphasise that, there is a linear dependency on the number and size of $\mathcal{O}_S$ w.r.t. merge processing time. The result quantifies that the n-ary merge is on average 4 (9) times faster than the balanced (ladder) binary merge, respectively.
This concludes that using n-ary rather than binary methods is more valuable and effective when the number of ontologies gets higher. For example, in $d_3$ with 3 source ontologies, n-ary is 3 times faster than binary strategies, in $d_6$ with $n=7$, n-ary is 4 times faster than both binary approaches, and in $d_{12}$ with $n=56$, it is 31 times faster than the binary ladder. 

Overall our results show that the n-ary strategy achieves comparable results in terms of quality but outperforms binary approaches in terms of runtime and complexity.

\section{State of the Art} \label{sec:relatedWork} 
Merging strategies basically have been divided into two main categories~\cite{batini1986comparative}: \textit{``binary''} and \textit{``n-ary''}. The \textit{binary} approach allows the merging of two ontologies at a time, while the \textit{n-ary} strategy merges $n$ ontologies ($n>2$) in a single step. To deal with merging more than two ontologies, the \textit{binary} strategy has a quadratic complexity of merging process in terms of involved ontologies. However, in the \textit{n-ary} strategy, the number of merging steps is minimized. Most methodologies in the literature, such as~\cite{OM,creado,HSSM,granualr,atom,OIMSM} agree on adopting a \textit{binary} strategy 
due to the simplicity of the search space. Applying a series of binary merges to more than two ontologies is not sufficiently scalable and viable for a large number of ontologies~\cite{rahm2016case}. The existing n-ary approaches~\cite{omersec,porsche} deal with merging multiple ontologies in a single step, however, each of these systems suffers certain drawbacks. In~\cite{porsche}, the final merge result depends on the order in which the source tree-structured XML schemas are matched and merged. 
In~\cite{omersec}, 
the experimental tests were carried out on a few small source ontologies, only. 
Porsche~\cite{porsche} does not target ontologies but XML schemas, and the code of OmerSec~\cite{omersec} is not available. Thus, we were not able to perform the comparison.
Despite the efforts of these research studies, developing an efficient, scalable n-ary method has not been practically applied and still is one of the crucial challenges. 

\par Merging ontologies either in a binary or n-ary approach can be categorized into two different strategies: \textit{``one-level merge''} and \textit{``two-level merge''}. In the latter one, an intermediate merge result is produced at the first level. Then, in the second phase, the intermediate result is refined to generate a final merge result. In contrast, \textit{one-level merge} approach~\cite{creado,HSSM} creates the merge result in one incrementally processing step by considering the effect of the previous combined entities. In the \textit{two-level merge} approaches~\cite{OM,omersec,granualr,atom,porsche,OIMSM}, a set of refinements is carried on the intermediated result. For instance, applying a set of GMRs in ATOM~\cite{atom}, utilizing granular processing in GCBOM~\cite{granualr}, and considering source ontologies' restrictions in OM~\cite{OM}. The output of the second level is generally called \textit{merged ontology}. Whereas, the outcome of the first level comes under different names, such as an \textit{integrated concept graph} in ATOM~\cite{atom}, \textit{network-based	knowledge model} in OIM-SM~\cite{OIMSM}, or \textit{intermediate schema} in PORSCHE~\cite{porsche}.

\section{Conclusion} \label{sec:conclusion}
Ontology merging is frequently needed. Existing approaches scale rather poorly to the merging of multiple ontologies. We proposed the n-ary multiple ontologies merging approach $\mathcal{C}$\textit{o}$\mathcal{M}$\textit{erger}, to overcome this issue. Efficiency is achieved by breaking  processing of $n$ ontologies into merging of $k$ blocks, with a minor overhead in the dividing process. 
The tool, that we built for this purpose, can provide a parameterizable merge platform allowing users  to influence the merge result interactively. 
Our future research agenda are possible strategies for such user interaction, adapting our approach to merging data on the schema-level of Linked Open Data (LOD) scenarios, and taking advantage of the parallelization potential of the approach. 

\section*{Acknowledgments}
S. Babalou is supported by a scholarship from German Academic Exchange Service (DAAD).
\setlength{\parskip}{-2pt} 
\bibliographystyle{ieeetr} 
\bibliography{HolisticAlg}

\newpage
\section*{Appendix}
Table~\ref{table:dataset1} shows the dataset statistics. 
The number of source ontologies $n$ in each dataset  is shown in \textit{column 2}. The source ontologies' name and the axioms' size $Sig(\mathcal{O}_S)$ are shown in \textit{columns 3-4}.
The number of corresponding classes $\mathcal{M}^{(')}_{|C|}$ and properties $\mathcal{M}^{(')}_{|P|}$ for imperfect and perfect mappings
has been presented in \textit{columns 5-8}, respectively.
\begin{table}[!b]
	\caption{Dataset statistics: number, name, and size of source ontologies with their imperfect and perfect mappings' size.}
	\centering
	\resizebox{\textwidth}{!} {
		\begin{tabular}{ | c | c | c | c || c | c || c | c |   }
			\hline
			\rowcolor{gray!15}
			id & $n$ & Source ontologies $\mathcal{O}_S$ & $|Sig(\mathcal{O}_S)|$&$\mathcal{M'}_{|C|}$ &$\mathcal{M'}_{|P|}$ &$\mathcal{M}_{|C|}$ &$\mathcal{M}_{|P|}$ \\ [0.5ex]
			\Xhline{3\arrayrulewidth}
			\cellcolor{gray!15}$d_1$ & 2 & cmt $|$ conference & 318 $|$ 408&7 & 2&11&3 \\ \hline
			\cellcolor{gray!15}$d_2$ & 2 & ekaw $|$ sigkdd & 341 $|$ 193& 8&1 & 11& 0 \\ \hline
			\cellcolor{gray!15}$d_3$ & 3 & cmt $|$ conference $|$ confOf & - $|$ - $|$ 335 & 14&5&22&11 \\ \hline
			\cellcolor{gray!15}$d_4$ & 4 & conference $|$ confOf $|$ edas $|$ ekaw  & - $|$ - $|$ 903 $|$ - & 33 & 14& 40& 13 \\ \hline
			\cellcolor{gray!15}$d_5$ & 4 & cmt $|$ ekaw $|$ iasted $|$ sigkdd  & - $|$ - $|$ 539 $|$ - & 29&8&33&5 \\ \hline
			\cellcolor{gray!15}&\multirow{2}{*}{7} &  cmt $|$ conference $|$ confOf $|$ edas & \multirow{2}{*}{- $|$ - $|$ - $|$ - $|$ - $|$ - $|$ -} & \multirow{2}{*}{57}&\multirow{2}{*}{25}&\multirow{2}{*}{68}&\multirow{2}{*}{27}\\
			\multirow{-2}{*}{\cellcolor{gray!15}{$d_6$}}& & ekaw $|$ iasted $|$ sigkdd &  &&&&\\ \hline
			\cellcolor{gray!15}$d_7$ & 2 & human $|$ mouse & 30364 $|$ 11043 &1175&2&1490&0\\ \hline
			\cellcolor{gray!15}$d_8$ & 2 & FMA\_samll $|$ NCI\_small  & 16690 $|$ 2472 & 2472&0&2480&0 \\ \hline
			\cellcolor{gray!15} & \multirow{2}{*}{7} & AHSO $|$ BNO $|$ CABRO   &  341 $|$ 281 $|$ 175 $|$ 741    &\multirow{2}{*}{6}  &\multirow{2}{*}{0}& \multirow{2}{*}{7}&\multirow{2}{*}{0} \\
			\multirow{-2}{*}{\cellcolor{gray!15}{$d_{9}$}}& & EPILONT $|$ RAO $|$ RNPRIO $|$ RO & 1401 $|$ 219 $|$ 2274 &&&& \\ \hline		
			\cellcolor{gray!15} & \multirow{4}{*}{17} & ADAR $|$ AHSO $|$ AO $|$ BNO  & 17857 $|$ - $|$ 872 $|$ - $|$ -  & 			\multirow{4}{*}{218}&\multirow{4}{*}{48} &\multirow{4}{*}{147}&\multirow{4}{*}{0}\\
			\cellcolor{gray!15}&&CABRO $|$ CSSO $|$ CWD $|$EPILONT &3472 $|$ 134 $|$ - $|$ 9438 $|$ 4665  &&&& \\
			\cellcolor{gray!15}& &ICO $|$ IDO $|$ NEOMARK3 $|$ NPI $|$ OF  & 2513 $|$ 401 $|$ 5007 $|$ 637  &&&& \\
			\multirow{-4}{*}{\cellcolor{gray!15}{$d_{10}$}}& & PCAO $|$ PHARE $|$ PSO $|$ RAO & 2169 $|$ 1879 $|$ 1401 & &&& \\ \hline			
			\cellcolor{gray!15}$d_{11}$ & 19 & $d_9$ and $d_{10}$'s ontologies& - & 219 &48&  155&0 \\ \hline		
			\cellcolor{gray!15} & \multirow{9}{*}{55} &$d_9$ and $d_{10}$'s ontologies and &1920 $|$ 2506 $|$ 572 $|$ 575  &\multirow{9}{*}{967} &\multirow{9}{*}{172}&\multirow{9}{*}{676}&\multirow{9}{*}{0}\\
			\cellcolor{gray!15}&&AEO $|$ AHOL $|$ AMINO-ACID $|$ BFO  &2619 $|$ 3020 $|$ 1993 $|$ 714 &&&& \\
			\cellcolor{gray!15}&& BHO $|$ BMT $|$ BOF $|$ BP $|$ BSPO  &1969 $|$ 2322 $|$ 591 $|$ 21144 &&&& \\
			\cellcolor{gray!15}&&BT $|$ CKDO $|$ CMPO $|$ CN $|$  DIAB & 1842 $|$ 5885 $|$ 446 $|$ 4617 &&&& \\
			\cellcolor{gray!15}&& DIAGONT $|$ EOL $|$ FBbi $|$ GDCO  &5383 $|$ 682 $|$ 479 $|$  2736  &&&& \\
			\cellcolor{gray!15}& &GFO $|$ GRO $|$ HIO $|$ INO $|$ LHN   &5392 $|$ 3769 $|$ 1828 $|$ 3280 & &&& \\
			\cellcolor{gray!15}& &MCBCC $|$ MEDO $|$ OBIWS $|$ ONLIRA   &355 $|$ 2241 $|$ 548 $|$ 3996 $|$ 551 &&&& \\
			\cellcolor{gray!15}& &OPB $|$ PEO $|$ PPO $|$  REPO $|$ RNPRIO  & 5401 $|$ 697 $|$ 219 $|$ 1726   &&&&\\
			\multirow{-9}{*}{\cellcolor{gray!15}{$d_{12}$}}&&ROS $|$ SHR  $|$ UNITSONT $|$ UO $|$ VIVO & 825 $|$ 340 $|$ 3629 $|$ 4862 &&&& \\ \hline
	\end{tabular}}
	\label{table:dataset1}
	\vspace{-4mm}%
\end{table}

\bigskip
Table~\ref{table:stOAEI} and Table~\ref{table:stBioportal} show the characteristics of the merged ontologies on the OAEI and BioPortal datasets, respectively. Parts of these two tables have already shown in Fig. 3 and Fig. 4 in the main manuscript. 

\begin{table}[bt]
	\caption{Characteristics of the n-ary merged result- OAEI dataset.}
	\centering
	\resizebox{9cm}{!} {
		\begin{tabular}{ | c | c | c | c | c | c | c | c | c | c | c | c | c |   }
			\hline
			\cellcolor{gray!15}& \multicolumn{7}{c|}{\cellcolor{gray!15}{Integrity}} &\multicolumn{3}{c|}{\cellcolor{gray!15}{Model Pr.}} &\multicolumn{2}{c|}{\cellcolor{gray!15}{Merge Pr.}}  \\ 
			[0.4ex] \cline{2-13} 
			\omit \vrule height.4pt\textcolor{gray!15}{\leaders\vrule\hfil}\vrule \cr 
			\cellcolor{gray!15}& \multicolumn{3}{c|}{\cellcolor{gray!15}{Compactness}}&\multicolumn{4}{c|}{\cellcolor{gray!15}{Coverage}}&\cellcolor{gray!15} & \cellcolor{gray!15} &\cellcolor{gray!15}&\cellcolor{gray!15} 
			& \cellcolor{gray!15}\\ 
			\cline{2-8}	
			\omit \vrule height.4pt\textcolor{gray!15}{\leaders\vrule\hfil}\vrule \cr 
			\rowcolor{gray!15}
			\multirow{-3}{*}{id}&$C$&$P$&$I$&$C$&$P$&$I$&$str$&\multirow{-2}{*}{$\ |on|\ $}&\multirow{-2}{*}{$|C_u|$}& \multirow{-2}{*}{$|cyc|$}&\multirow{-2}{*}{$|R_L|$}&\multirow{-2}{*}{$|R_G|$}
			\\ \Xhline{3\arrayrulewidth}
			\cellcolor{gray!15}$d_{1}V_1$    & 77        & 120          & 0            & 1               & 1                  & -                  & 0                      & 0     & 0                  & 0            & 14                          & 8                         \\ \hline 
			\cellcolor{gray!15}$d_{1}V_2$    & 77        & 120          & 0            & 1               & 1                  & -                  & 0                      & 0     & 0                  & 0            & -                           & 12                        \\ \hline 
			\cellcolor{gray!15}$d_{1}V_3$    & 76        & 120          & 0            & 0.97            & 1                  & -                  & 0                      & 4     & 6                  & 1            & -                           & 0                         \\ \hline 
			\cellcolor{gray!15}$d_{1}V_4$    & 82        & 121          & 0            & 1               & 1                  & -                  & 0                      & 0     & 0                  & 0            & 8                           & 12                        \\ \hline 
			\cellcolor{gray!15}$d_{1}V_5$    & 82        & 121          & 0            & 1               & 1                  & -                  & 0                      & 0     & 0                  & 0            & -                           & 13                        \\ \hline 
			\cellcolor{gray!15}$d_{1}V_6$    & 81        & 121          & 0            & 0.98            & 1                  & -                  & 0                      & 4     & 8                  & 0            & -                           & 0                         \\ \Xhline{3\arrayrulewidth}
			\cellcolor{gray!15}$d_{2}V_1$    & 112       & 61           & 0            & 1               & 1                  & -                  & 0                      & 0     & 0                  & 0            & 17                          & 1                         \\ \hline 
			\cellcolor{gray!15}$d_{2}V_2$    & 112       & 61           & 0            & 1               & 1                  & -                  & 0                      & 0     & 0                  & 0            & -                           & 1                         \\ \hline 
			\cellcolor{gray!15}$d_{2}V_3$    & 112       & 61           & 0            & 1               & 1                  & -                  & 0                      & 0     & 1                  & 0            & -                           & 0                         \\ \hline 
			\cellcolor{gray!15}$d_{2}V_4$    & 115       & 60           & 0            & 1               & 1                  & -                  & 0                      & 0     & 0                  & 0            & 18                          & 3                         \\ \hline 
			\cellcolor{gray!15}$d_{2}V_5$    & 115       & 60           & 0            & 1               & 1                  & -                  & 0                      & 0     & 0                  & 0            & -                           & 4                         \\ \hline 
			\cellcolor{gray!15}$d_{2}V_6$    & 115       & 60           & 0            & 1               & 1                  & -                  & 0                      & 2     & 2                  & 0            & -                           & 0                         \\ \Xhline{3\arrayrulewidth}
			\cellcolor{gray!15}$d_{3}V_1$    & 98        & 147          & 0            & 0.98            & 1                  & -                  & 0                      & 0     & 0                  & 0            & 21                          & 11                        \\ \hline 
			\cellcolor{gray!15}$d_{3}V_2$    & 98        & 147          & 0            & 0.98            & 1                  & -                  & 0                      & 0     & 0                  & 0            & -                           & 16                        \\ \hline 
			\cellcolor{gray!15}$d_{3}V_3$    & 97        & 147          & 0            & 0.97            & 1                  & -                  & 0                      & 6     & 7                  & 1            & -                           & 0                         \\ \hline 
			\cellcolor{gray!15}$d_{3}V_4$    & 109       & 154          & 0            & 0.98            & 1                  & -                  & 0                      & 0     & 0                  & 0            & 14                          & 15                        \\ \hline 
			\cellcolor{gray!15}$d_{3}V_5$    & 109       & 154          & 0            & 0.98            & 1                  & -                  & 0                      & 0     & 0                  & 0            & -                           & 17                        \\ \hline 
			\cellcolor{gray!15}$d_{3}V_6$    & 108       & 154          & 0            & 0.97            & 1                  & -                  & 0                      & 6     & 9                  & 0            & -                           & 0                         \\ \Xhline{3\arrayrulewidth}
			\cellcolor{gray!15}$d_{4}V_1$    & 202       & 167          & 114          & 0.99            & 1                  & 1                  & 0                      & 0     & 0                  & 0            & 30                          & 12                        \\ \hline 
			\cellcolor{gray!15}$d_{4}V_2$    & 201       & 167          & 114          & 0.99            & 1                  & 1                  & 0                      & 0     & 0                  & 0            & -                           & 18                        \\ \hline 
			\cellcolor{gray!15}$d_{4}V_3$    & 199       & 167          & 114          & 0.98            & 1                  & 1                  & 0                      & 6     & 8                  & 1            & -                           & 0                         \\ \hline 
			\cellcolor{gray!15}$d_{4}V_4$    & 226       & 165          & 114          & 0.99            & 0.99               & 1                  & 0                      & 0     & 0                  & 0            & 26                          & 18                        \\ \hline 
			\cellcolor{gray!15}$d_{4}V_5$    & 226       & 165          & 114          & 0.99            & 0.99               & 1                  & 0                      & 0     & 0                  & 0            & -                           & 21                        \\ \hline 
			\cellcolor{gray!15}$d_{4}V_6$    & 224       & 165          & 114          & 0.98            & 0.99               & 1                  & 0                      & 9     & 9                  & 0            & -                           & 0                         \\ \Xhline{3\arrayrulewidth}
			\cellcolor{gray!15}$d_{5}V_1$    & 248       & 156          & 4            & 0.99            & 1                  & 1                  & 0                      & 0     & 0                  & 0            & 61                          & 7                         \\ \hline 
			\cellcolor{gray!15}	$d_{5}V_2$    & 247       & 156          & 4            & 0.99            & 1                  & 1                  & 0                      & 0     & 0                  & 0            & -                           & 9                         \\ \hline 
			\cellcolor{gray!15}$d_{5}V_3$    & 246       & 156          & 4            & 0.98            & 1                  & 1                  & 0                      & 3     & 5                  & 0            & -                           & 0                         \\ \hline 
			\cellcolor{gray!15}$d_{5}V_4$    & 253       & 155          & 4            & 0.99            & 1.01               & 1                  & 0                      & 0     & 0                  & 0            & 64                          & 13                        \\ \hline 
			\cellcolor{gray!15}$d_{5}V_5$    & 253       & 154          & 4            & 0.99            & 1                  & 1                  & 0                      & 0     & 0                  & 0            & -                           & 17                        \\ \hline 
			\cellcolor{gray!15}	$d_{5}V_6$    & 251       & 153          & 4            & 0.98            & 1                  & 1                  & 1                      & 9     & 5                  & 0            & -                           & 0                         \\ \Xhline{3\arrayrulewidth}
			\cellcolor{gray!15}$d_{6}V_1$    & 338       & 274          & 118          & 0.99            & 1.05               & 1                  & 0                      & 0     & 0                  & 0            & 71                          & 21                        \\ \hline 
			\cellcolor{gray!15}$d_{6}V_2$    & 336       & 274          & 118          & 0.98            & 1.05               & 1                  & 0                      & 0     & 0                  & 0            & -                           & 29                        \\ \hline 
			\cellcolor{gray!15}$d_{6}V_3$    & 334       & 274          & 118          & 0.98            & 1.05               & 1                  & 0                      & 14    & 8                  & 4            & -                           & 0                         \\ \hline 
			\cellcolor{gray!15}$d_{6}V_4$    & 390       & 283          & 118          & 0.99            & 1.03               & 1                  & 0                      & 0     & 0                  & 0            & 96                          & 34                        \\ \hline 
			\cellcolor{gray!15}$d_{6}V_5$    & 390       & 281          & 118          & 0.99            & 1.02               & 1                  & 0                      & 0     & 0                  & 0            & -                           & 41                        \\ \hline 
			\cellcolor{gray!15}$d_{6}V_6$    & 386       & 280          & 118          & 0.98            & 1.01               & 1                  & 1                      & 24    & 11                 & 0            & -                           & 0                         \\ \Xhline{3\arrayrulewidth}
			\cellcolor{gray!15}$d_{7}V_1$    & 4526      & 4            & 0            & 1               & 1                  & -                  & 0                      & 0     & 0                  & 0            & -                           & 9                         \\ \hline 
			\cellcolor{gray!15}$d_{7}V_2$    & 4526      & 4            & 0            & 1               & 1                  & -                  & 0                      & 0     & 0                  & 0            & -                           & 9                         \\ \hline 
			\cellcolor{gray!15}$d_{7}V_3$    & 4526      & 4            & 0            & 1               & 1                  & -                  & 0                      & 0     & 7                  & 2            & -                           & 0                         \\ \hline 
			\cellcolor{gray!15}$d_{7}V_4$    & 4873      & 2            & 0            & 1               & 1                  & -                  & 0                      & 0     & 0                  & 0            & -                           & 7                         \\ \hline 
			\cellcolor{gray!15}$d_{7}V_5$    & 4873      & 2            & 0            & 1               & 1                  & -                  & 0                      & 0     & 0                  & 0            & -                           & 7                         \\ \hline 
			\cellcolor{gray!15}$d_{7}V_6$    & 4873      & 2            & 0            & 1               & 1                  & -                  & 0                      & 0     & 7                  & 0            & -                           & 0                         \\ \Xhline{3\arrayrulewidth}
			\cellcolor{gray!15}$d_{8}V_1$    & 7290      & 87           & 0            & 1               & 1                  & -                  & 0                      & 0     & 0                  & 0            & -                           & 1949                      \\ \hline 
			\cellcolor{gray!15}$d_{8}V_2$    & 7290      & 87           & 0            & 1               & 1                  & -                  & 0                      & 0     & 0                  & 0            & -                           & 1949                      \\ \hline 
			\cellcolor{gray!15}$d_{8}V_3$    & 7285      & 87           & 0            & 1               & 1                  & -                  & 4                      & 0     & 1916               & 29           & -                           & 0                         \\ \hline 
			\cellcolor{gray!15}$d_{8}V_4$    & 7721      & 87           & 0            & 1               & 1                  & -                  & 0                      & 0     & 0                  & 0            & -                           & 1925                      \\ \hline 
			\cellcolor{gray!15}$d_{8}V_5$    & 7721      & 87           & 0            & 1               & 1                  & -                  & 0                      & 0     & 0                  & 0            & -                           & 1925                      \\ \hline 
			\cellcolor{gray!15}$d_{8}V_6$    & 7712      & 87           & 0            & 1               & 1                  & -                  & 9                      & 0     & 1916               & 0            & -                           & 0                        	
			\\ \hline                     
	\end{tabular}}
	\label{table:stOAEI} 
	\vspace{-4mm}%
\end{table}

\begin{table}[bt]
	\caption{Characteristics of the n-ary merged result- BioPortal dataset.}
	\centering
	\resizebox{\textwidth}{!} {
		\begin{tabular}{ | c | c | c | c | c | c | c | c | c | c | c | c | c |   }
			\hline
			\cellcolor{gray!15}& \multicolumn{7}{c|}{\cellcolor{gray!15}{Integrity}} &\multicolumn{3}{c|}{\cellcolor{gray!15}{Model Pr.}} &\multicolumn{2}{c|}{\cellcolor{gray!15}{Merge Pr.}}  \\ 
			[0.4ex] \cline{2-13} 
			\omit \vrule height.4pt\textcolor{gray!15}{\leaders\vrule\hfil}\vrule \cr 
			\cellcolor{gray!15}& \multicolumn{3}{c|}{\cellcolor{gray!15}{Compactness}}&\multicolumn{4}{c|}{\cellcolor{gray!15}{Coverage}}&\cellcolor{gray!15} & \cellcolor{gray!15} &\cellcolor{gray!15}&\cellcolor{gray!15} 
			& \cellcolor{gray!15}\\ 
			\cline{2-8}\omit \vrule height.4pt\textcolor{gray!15}{\leaders\vrule\hfil}\vrule \cr 
			\rowcolor{gray!15}
			\multirow{-3}{*}{id}&$C$&$P$&$I$&$C$&$P$&$I$&$str$&\multirow{-2}{*}{$\ |on|\ $}&\multirow{-2}{*}{$|C_u|$}& \multirow{-2}{*}{$|cyc|$}&\multirow{-2}{*}{$|R_L|$}&\multirow{-2}{*}{$|R_G|$}
			\\ \Xhline{3\arrayrulewidth}
			\cellcolor{gray!15}$d_{9}V_1$    & 1145      & 101          & 31           & 1               & 0.72               & 1                  & 0                      & 0     & 0                  & 0            & 8                           & 17                        \\ \hline 
			\cellcolor{gray!15}$d_{9}V_2$    & 1145      & 101          & 31           & 1               & 0.72               & 1                  & 0                      & 0     & 0                  & 0            & -                           & 17                        \\ \hline 
			\cellcolor{gray!15}$d_{9}V_3$    & 1144      & 101          & 31           & 1               & 0.72               & 1                  & 0                      & 14    & 2                  & 0            & -                           & 0                         \\ \hline 
			\cellcolor{gray!15}$d_{9}V_4$    & 1146      & 101          & 31           & 1               & 0.72               & 1                  & 0                      & 0     & 0                  & 0            & 0                           & 17                        \\ \hline 
			\cellcolor{gray!15}$d_{9}V_5$    & 1146      & 101          & 31           & 1               & 0.72               & 1                  & 0                      & 0     & 0                  & 0            & -                           & 17                        \\ \hline 
			\cellcolor{gray!15}$d_{9}V_6$    & 1145      & 101          & 31           & 1               & 0.72               & 1                  & 0                      & 14    & 2                  & 0            & -                           & 0                         \\ \Xhline{3\arrayrulewidth}
			\cellcolor{gray!15}$d_{10}V_1$   & 5042      & 2197         & 843          & 0.99            & 0.9                & 0.93               & 0                      & 0     & 0                  & 0            & 41                          & 106                       \\ \hline 
			\cellcolor{gray!15}$d_{10}V_2$   & 5042      & 2197         & 843          & 0.99            & 0.9                & 0.93               & 0                      & 0     & 0                  & 0            & -                           & 116                       \\ \hline 
			\cellcolor{gray!15}$d_{10}V_3$   & 5018      & 2197         & 843          & 0.99            & 0.9                & 0.93               & 71                     & 21    & 5                  & 2            & -                           & 0                         \\ \hline 
			\cellcolor{gray!15}$d_{10}V_4$   & 4957      & 2394         & 882          & 1               & 1.01               & 0.98               & 0                      & 0     & 0                  & 0            & -                           & 143                       \\ \hline 
			\cellcolor{gray!15}$d_{10}V_5$   & 4957      & 2394         & 882          & 1               & 1.01               & 0.98               & 0                      & 0     & 0                  & 0            & -                           & 143                       \\ \hline 
			\cellcolor{gray!15}$d_{10}V_6$   & 4928      & 2378         & 882          & 0.99            & 1                  & 0.98               & 87                     & 22    & 7                  & 3            & -                           & 0                         \\ \Xhline{3\arrayrulewidth}
			\cellcolor{gray!15}$d_{11}V_1$   & 5564      & 2245         & 870          & 0.99            & 0.89               & 0.94               & 0                      & 0     & 0                  & 0            & 40                          & 110                       \\ \hline 
			\cellcolor{gray!15}$d_{11}V_2$   & 5564      & 2245         & 870          & 0.99            & 0.89               & 0.94               & 0                      & 0     & 0                  & 0            & -                           & 120                       \\ \hline 
			\cellcolor{gray!15}$d_{11}V_3$   & 5539      & 2245         & 870          & 0.99            & 0.89               & 0.94               & 75                     & 21    & 5                  & 2            & -                           & 0                         \\ \hline 
			\cellcolor{gray!15}$d_{11}V_4$   & 5490      & 2469         & 909          & 1               & 1.01               & 0.98               & 0                      & 0     & 0                  & 0            & -                           & 143                       \\ \hline 
			\cellcolor{gray!15}$d_{11}V_5$   & 5490      & 2469         & 909          & 1               & 1.01               & 0.98               & 0                      & 0     & 0                  & 0            & -                           & 143                       \\ \hline 
			\cellcolor{gray!15}$d_{11}V_6$   & 5461      & 2453         & 909          & 0.99            & 1                  & 0.98               & 87                     & 22    & 7                  & 3            & -                           & 0                         \\ \Xhline{3\arrayrulewidth}
			\cellcolor{gray!15}$d_{12}V_1$   & 15822     & 3818         & 1262         & 0.98            & 0.95               & 0.96               & 0                      & 0     & 0                  & 0            & -                           & 524                       \\ \hline 
			\cellcolor{gray!15}$d_{12}V_2$   & 15822     & 3818         & 1262         & 0.98            & 0.95               & 0.96               & 0                      & 0     & 0                  & 0            & -                           & 524                       \\ \hline 
			\cellcolor{gray!15}$d_{12}V_3$   & 15729     & 3818         & 1262         & 0.98            & 0.95               & 0.96               & 176                    & 44    & 252                & 4            & -                           & 0                         \\ \hline 
			\cellcolor{gray!15}$d_{12}V_4$   & 15080     & 3589         & 1262         & 0.99            & 1.03               & 0.96               & 0                      & 0     & 0                  & 0            & -                           & 977                       \\ \hline 
			\cellcolor{gray!15}$d_{12}V_5$   & 15080     & 3589         & 1262         & 0.99            & 1.03               & 0.96               & 0                      & 0     & 0                  & 0            & -                           & 977                       \\ \hline 
			\cellcolor{gray!15}$d_{12}V_6$   & 14944     & 3540         & 1262         & 0.98            & 1.01               & 0.96               & 579                    & 85    & 229                & 13           & -                           & 0                  \\ \hline                     
	\end{tabular}}
	\label{table:stBioportal} 
	\vspace{-4mm}%
\end{table}

\bigskip
Table~\ref{table:binHolFull1}, Table~\ref{table:binHolFull2}, and Table~\ref{table:binHolFull3} compare the n-ary, balanced, and ladder merge strategies for different versions of merged ontology based on the test setting of Section 3.2 in the main manuscript. Table~\ref{table:binHolFull1} shows the results when both local and global refinements are applied. Table~\ref{table:binHolFull2} shows the results when only global refinements are applied, which part of it has already been presented in the main manuscript. Table~\ref{table:binHolFull3} shows the results when no refinement is applied. In all tables, we show the results for those datasets which have more than two source ontologies.

\begin{table}[bt!]
	\caption{Comparing n-ary ($V_4$), balanced ($V_7$), and ladder($V_{10}$)  merge strategies.}
	\centering
	\resizebox{\textwidth}{!} {
		\begin{tabular}{| c | c | c | c | c | c | c | c | c | c | }
			\hline
			\rowcolor{gray!15}
			id & Method& $|C|$ & $|P|$ & $|I|$ & $|Cor|$ & $|tr|$ & $|R_L|$ & $|R_G|$ & $|Mer.|$ \\ \Xhline{3\arrayrulewidth} 
			\cellcolor{gray!15}    & N-ary    & 109       & 154          & 0            & 19   & 374               & 14                          & 15                        & 1    \\ \cline{2-10}
			\cellcolor{gray!15}   & Balanced & 110       & 154          & 0            & 22   & 566               & 37                          & 16                        & 2    \\ \cline{2-10}
			\multirow{-3}{*}{\cellcolor{gray!15}{$d_3$}}    & Ladder   & 110       & 154          & 0            & 22   & 566               & 38                          & 16                        & 2    \\ \Xhline{3\arrayrulewidth} 
			\cellcolor{gray!15} & N-ary    & 226       & 165          & 114          & 47   & 790               & 26                          & 18                        & 1    \\ \cline{2-10}
			\cellcolor{gray!15}    & Balanced & 226       & 166          & 114          & 65   & 1270              & 26                          & 27                        & 3    \\ \cline{2-10}
			\multirow{-3}{*}{\cellcolor{gray!15}{$d_4$}}    & Ladder   & 226       & 167          & 114          & 64   & 1339              & 122                         & 21                        & 3    \\\Xhline{3\arrayrulewidth} 
			\cellcolor{gray!15}    & N-ary    & 253       & 155          & 4            & 37   & 527               & 64                          & 13                        & 1    \\ \cline{2-10}
			\cellcolor{gray!15}  & Balanced & 254       & 156          & 4            & 47   & 863               & 162                         & 16                        & 3    \\ \cline{2-10}
			\multirow{-3}{*}{\cellcolor{gray!15}{$d_5$}}    & Ladder   & 255       & 156          & 4            & 47   & 997               & 214                         & 15                        & 3    \\ \Xhline{3\arrayrulewidth} 
			\cellcolor{gray!15}   & N-ary    & 390       & 283          & 118          & 82   & 1310              & 96                          & 34                        & 1    \\ \cline{2-10}
			\cellcolor{gray!15}   & Balanced & 395       & 285          & 118          & 127  & 2785              & 398                         & 37                        & 6    \\ \cline{2-10}
			\multirow{-3}{*}{\cellcolor{gray!15}{$d_6$}}    & Ladder   & 396       & 285          & 118          & 128  & 3406              & 521                         & 42                        & 6    \\ \Xhline{3\arrayrulewidth} 
			\cellcolor{gray!15}    & N-ary    & 1146      & 101          & 31           & 6    & 278               & 0                           & 17                        & 1    \\ \cline{2-10}
			\cellcolor{gray!15}   & Balanced & 1146      & 101          & 31           & 6    & 360               & 2                           & 17                        & 6    \\ \cline{2-10}
			\multirow{-3}{*}{\cellcolor{gray!15}{$d_9$}}    & Ladder   & 1146      & 139          & 31           & 6    & 483               & 0                           & 17                        & 6    \\ \Xhline{3\arrayrulewidth} 
			\cellcolor{gray!15} & N-ary    & 4957      & 2394         & 882          & 266  & 6949              & -                           & 143                       & 1    \\ \cline{2-10}
			\cellcolor{gray!15} & Balanced & 4984      & 2194         & 843          & 368  & 15790             & 272                         & 147                       & 16   \\ \cline{2-10}
			\multirow{-3}{*}{\cellcolor{gray!15}{$d_{10}$}} & Ladder   & 5003      & 2195         & 843          & 351  & 31415             & 949                         & 173                       & 16   \\ \Xhline{3\arrayrulewidth} 
			\cellcolor{gray!15} & N-ary    & 5490      & 2469         & 909          & 267  & 6960              & -                           & 143                       & 1    \\ \cline{2-10}
			\cellcolor{gray!15} & Balanced & 5142      & 2248         & 870          & 730  & 19024             & 102                         & 155                       & 18   \\ \cline{2-10}
			\multirow{-3}{*}{\cellcolor{gray!15}{$d_{11}$}} & Ladder   & 5510      & 2266         & 870          & 387  & 35489             & 951                         & 178                       & 18   \\ \Xhline{3\arrayrulewidth} 
			\cellcolor{gray!15} & N-ary    & 15080     & 3589         & 1262         & 1139 & 30035             & -                           & 977                       & 1    \\ \cline{2-10}
			\cellcolor{gray!15} & Balanced & 15285     & 3227         & 1175         & 4136 & 82041             & 2095                        & 1517                      & 54   \\ \cline{2-10}
			\multirow{-3}{*}{\cellcolor{gray!15}{$d_{12}$}} & Ladder   & 15384     & 3398         & 1175         & 2156 & 153871            & 3831                        & 1541                      & 54  \\ \hline
	\end{tabular}}
	\label{table:binHolFull1} 
	\vspace{-4mm}%
\end{table}

\begin{table}[bt!]
	\caption{Comparing n-ary($V_5$),balanced ($V_{8}$), and ladder ($V_{11}$) merge strategies.}
	\centering
	\resizebox{\textwidth}{!} {
		\begin{tabular}{| c | c | c | c | c | c | c | c | c | }
			\hline
			\rowcolor{gray!15}
			id & Method& $|C|$ & $|P|$ & $|I|$ & $|Cor|$ & $|tr|$ & $|R_G|$ & $|Mer.|$ \\ \Xhline{3\arrayrulewidth} 
			\cellcolor{gray!15}   & N-ary    & 109       & 154          & 0            & 19   & 374               & 17                        & 1    \\ \cline{2-9}
			\cellcolor{gray!15}   & Balanced & 110       & 154          & 0            & 22   & 566               & 18                        & 2    \\ \cline{2-9}
			\multirow{-3}{*}{\cellcolor{gray!15}{$d_3$}}    & Ladder   & 110       & 154          & 0            & 22   & 566               & 18                        & 2    \\  \Xhline{3\arrayrulewidth}
			\cellcolor{gray!15}  & N-ary    & 226       & 165          & 114          & 47   & 790               & 21                        & 1    \\ \cline{2-9}
			\cellcolor{gray!15}    & Balanced & 226       & 166          & 114          & 65   & 1270              & 27                        & 3    \\ \cline{2-9}
			\multirow{-3}{*}{\cellcolor{gray!15}{$d_4$}}    & Ladder   & 226       & 167          & 114          & 64   & 1339              & 23                        & 3    \\ \Xhline{3\arrayrulewidth}
			\cellcolor{gray!15}    & N-ary    & 253       & 154          & 4            & 37   & 527               & 17                        & 1    \\ \cline{2-9}
			\cellcolor{gray!15}   & Balanced & 254       & 154          & 4            & 47   & 865               & 19                        & 3    \\\cline{2-9}
			\multirow{-3}{*}{\cellcolor{gray!15}{$d_5$}}    & Ladder   & 254       & 154          & 4            & 47   & 994               & 19                        & 3    \\ \Xhline{3\arrayrulewidth}
			\cellcolor{gray!15}    & N-ary    & 390       & 281          & 118          & 82   & 1310              & 41                        & 1    \\ \cline{2-9}
			\cellcolor{gray!15}    & Balanced & 394       & 284          & 118          & 127  & 2791              & 48                        & 6    \\ \cline{2-9}
			\multirow{-3}{*}{\cellcolor{gray!15}{$d_6$}}    & Ladder   & 394       & 283          & 118          & 128  & 3462              & 51                        & 6    \\ \Xhline{3\arrayrulewidth}
			\cellcolor{gray!15}    & N-ary    & 1146      & 101          & 31           & 6    & 278               & 17                        & 1    \\ \cline{2-9}
			\cellcolor{gray!15}    & Balanced & 1146      & 101          & 31           & 6    & 360               & 17                        & 6    \\ \cline{2-9}
			\multirow{-3}{*}{\cellcolor{gray!15}{$d_9$}}    & Ladder   & 1146      & 139          & 31           & 6    & 483               & 17                        & 6    \\ \Xhline{3\arrayrulewidth}
			\cellcolor{gray!15} & N-ary    & 4957      & 2394         & 882          & 266  & 6949              & 143                       & 1    \\ \cline{2-9}
			\cellcolor{gray!15} & Balanced & 4982      & 2193         & 843          & 369  & 15816             & 166                       & 16   \\ \cline{2-9}
			\multirow{-3}{*}{\cellcolor{gray!15}{$d_{10}$}} & Ladder   & 5003      & 2193         & 843          & 351  & 31420             & 191                       & 16   \\ \Xhline{3\arrayrulewidth}
			\cellcolor{gray!15} & N-ary    & 5490      & 2469         & 909          & 267  & 6960              & 143                       & 1    \\ \cline{2-9}
			\cellcolor{gray!15} & Balanced & 5507      & 2246         & 870          & 388  & 16060             & 177                       & 18   \\ \cline{2-9}
			\multirow{-3}{*}{\cellcolor{gray!15}{$d_{11}$}} & Ladder   & 5511      & 2265         & 870          & 386  & 35480             & 196                       & 18   \\ \Xhline{3\arrayrulewidth}
			\cellcolor{gray!15} & N-ary    & 15080     & 3589         & 1262         & 1139 & 30035             & 977                       & 1    \\ \cline{2-9}
			\cellcolor{gray!15}& Balanced & 15450     & 3282         & 1175         & 2186 & 66344             & 1533                      & 54   \\ \cline{2-9}
			\multirow{-3}{*}{\cellcolor{gray!15}{$d_{12}$}} & Ladder   & 15384     & 3395         & 1175         & 2159 & 154002            & 1560                      & 54  \\ \hline
	\end{tabular}}
	\label{table:binHolFull2} 
	\vspace{-4mm}%
\end{table}

\begin{table}[bt!]
	\caption{Comparing n-ary($V_6$),balanced ($V_{9}$), and ladder ($V_{12}$) merge strategies.}
	\centering
	\resizebox{9cm}{!} {
		\begin{tabular}{| c | c | c | c | c | c | c | c |  }
			\hline
			\rowcolor{gray!15}
			id & Method& $|C|$ & $|P|$ & $|I|$ & $|Cor|$ & $|tr|$ & $|Mer.|$ \\ \Xhline{3\arrayrulewidth} 
			\cellcolor{gray!15}    & N-ary    & 108       & 154          & 0            & 19   & 374               & 1                     \\ \cline{2-8}
			\cellcolor{gray!15}    & Balanced & 109       & 154          & 0            & 22   & 567               & 2                     \\ \cline{2-8}
			\multirow{-3}{*}{\cellcolor{gray!15}{$d_3$}}    & Ladder   & 109       & 154          & 0            & 22   & 567               & 2                     \\ \Xhline{3\arrayrulewidth} 
			\cellcolor{gray!15}    & N-ary    & 224       & 165          & 114          & 47   & 790               & 1                     \\ \cline{2-8}
			\cellcolor{gray!15}  & Balanced & 224       & 166          & 114          & 65   & 1268              & 3                     \\ \cline{2-8}
			\multirow{-3}{*}{\cellcolor{gray!15}{$d_4$}}    & Ladder   & 224       & 167          & 114          & 64   & 1339              & 3                     \\ \Xhline{3\arrayrulewidth} 
			\cellcolor{gray!15}    & N-ary    & 251       & 153          & 4            & 37   & 527               & 1                     \\ \cline{2-8}
			\cellcolor{gray!15}    & Balanced & 252       & 153          & 4            & 47   & 868               & 3                     \\ \cline{2-8}
			\multirow{-3}{*}{\cellcolor{gray!15}{$d_5$}}    & Ladder   & 252       & 153          & 4            & 47   & 1001              & 3                     \\ \Xhline{3\arrayrulewidth} 
			\cellcolor{gray!15}    & N-ary    & 386       & 280          & 118          & 82   & 1310              & 1                     \\ \cline{2-8}
			\cellcolor{gray!15}    & Balanced & 389       & 283          & 118          & 127  & 2786              & 6                     \\ \cline{2-8}
			\multirow{-3}{*}{\cellcolor{gray!15}{$d_6$}}    & Ladder   & 390       & 282          & 118          & 128  & 3440              & 6                     \\ \Xhline{3\arrayrulewidth} 
			\cellcolor{gray!15}    & N-ary    & 1145      & 101          & 31           & 6    & 278               & 1                     \\ \cline{2-8}
			\cellcolor{gray!15}    & Balanced & 1145      & 101          & 31           & 6    & 372               & 6                     \\ \cline{2-8}
			\multirow{-3}{*}{\cellcolor{gray!15}{$d_9$}}    & Ladder   & 1146      & 139          & 31           & 6    & 513               & 6                     \\ \Xhline{3\arrayrulewidth} 
			\cellcolor{gray!15} & N-ary    & 4928      & 2378         & 882          & 266  & 6949              & 1                     \\ \cline{2-8}
			\cellcolor{gray!15} & Balanced & 4951      & 2177         & 843          & 365  & 15866             & 16                    \\ \cline{2-8}
			\multirow{-3}{*}{\cellcolor{gray!15}{$d_{10}$}} & Ladder   & 4970      & 2176         & 843          & 350  & 31733             & 16                    \\ \Xhline{3\arrayrulewidth} 
			\cellcolor{gray!15} & N-ary    & 5461      & 2453         & 909          & 267  & 6960              & 1                     \\ \cline{2-8}
			\cellcolor{gray!15} & Balanced & 5475      & 2231         & 870          & 386  & 15801             & 18                    \\ \cline{2-8}
			\multirow{-3}{*}{\cellcolor{gray!15}{$d_{11}$}} & Ladder   & 5474      & 2246         & 870          & 385  & 35742             & 18                    \\ \Xhline{3\arrayrulewidth} 
			\cellcolor{gray!15} & N-ary    & 14944     & 3540         & 1262         & 1139 & 30035             & 1                     \\ \cline{2-8}
			\cellcolor{gray!15} & Balanced & 14868     & 3225         & 1171         & 3617 & 80151             & 54                    \\ \cline{2-8}
			\multirow{-3}{*}{\cellcolor{gray!15}{$d_{12}$}} & Ladder   & 15111     & 3346         & 1175         & 2135 & 156293            & 54   	\\ \hline
	\end{tabular}}
	\label{table:binHolFull3} 
	\vspace{-4mm}%
\end{table}

\end{sloppypar}
\end{document}